\documentclass[twocolumn]{article}

\usepackage[T1]{fontenc}
\usepackage{lmodern}
\usepackage{dblfloatfix}
\usepackage[margin=1in]{geometry}
\usepackage{amsmath}
\usepackage{amssymb}
\usepackage{booktabs}
\usepackage{graphicx}
\usepackage{enumitem}
\usepackage{microtype}
\usepackage{listings}
\usepackage{csquotes}
\usepackage{algorithm}
\usepackage{algpseudocode}
\usepackage{tabularx}
\usepackage{tabularray}
\usepackage{tikz}
\usepackage{amsthm}
\usepackage[font=small,labelfont=bf]{caption}
\usepackage{fancyhdr}
\usepackage{ulem}   % for strikethrough
\usepackage{placeins}
\usepackage{array}
\usepackage{makecell}
\usepackage{pifont}
\usepackage{multirow}
\usepackage{hyphenat}
\usepackage{xurl}
\usepackage{titlesec}

\titleformat{\section}
{\large\bfseries\raggedright} {\thesection} {1em} {}

\usepackage[ backend=biber, bibstyle=authoryear, citestyle=authoryear,
  maxcitenames=2, giveninits=true, uniquename=init, doi=true, url=true,
  isbn=false ]{biblatex}

\usepackage[ colorlinks=true, linkcolor=blue, citecolor=blue, urlcolor=blue
]{hyperref}

\UseTblrLibrary{booktabs}
\usetikzlibrary{arrows.meta,positioning,decorations.pathreplacing,
shapes.geometric,calc}

\AtBeginBibliography{\emergencystretch=3em}
\DeclareDelimFormat{nameyeardelim}{\addcomma\space}

\newcommand{\full}{\checkmark}
\newcommand{\partialcov}{$\circ$}
\newcommand{\none}{--}

\newcommand{\cmark}{\ding{51}} \newcommand{\xmark}{\ding{55}}
\newcommand{\partialmark}{\(\circ\)}

\newcommand{\appsection}[2]{%
  \refstepcounter{section}%
  \phantomsection%
  \section*{Appendix \thesection: #1}%
  \markboth{Appendix \thesection: #1}{Appendix \thesection: #1}%
  \addcontentsline{toc}{section}{Appendix \thesection: #1}%
  \label{#2}%
}

\newcommand{\hfmodel}[1]{%
  \href{https://huggingface.co/#1}{\nolinkurl{#1}}%
}

\newcommand{\failmark}{[invalid]}

\newcommand{\aegitbranch}{main}
\newcommand{\aegitbase}{%
  https://github.com/victorlavrenko/answer-engineering/blob/\aegitbranch%
}
\newcommand{\ghfile}[1]{%
  \href{\aegitbase/#1}{\nolinkurl{#1}}%
}

\newcommand{\aehfrun}{run-20260428T134716Z}
\newcommand{\aehfbase}{%
  https://huggingface.co/datasets/lavrenko/answer-engineering/tree/main/\aehfrun%
}
\newcommand{\hfrunroot}{%
  \href{\aehfbase}{\nolinkurl{\aehfrun}}%
}
\newcommand{\hfdir}[1]{%
  \href{\aehfbase/#1}{\nolinkurl{\aehfrun/#1}}%
}

\newcommand{\aehfsamplesubrun}{subrun-007-trajectory-editing-orl-ssnhl-acute}
\newcommand{\hfsamplesubrun}{%
  \href{\aehfbase/\aehfsamplesubrun}{\nolinkurl{\aehfrun/\aehfsamplesubrun}}%
}
\newcommand{\hfsampledir}[1]{%
  \href{\aehfbase/\aehfsamplesubrun/#1}%
       {\nolinkurl{\aehfrun/\aehfsamplesubrun/#1}}%
}

\newcommand{\PaperEvalN}{1000}

\newcommand{\SSNHLBaselineAcceptedPct}{54.5\%}

\newcommand{\SSNHLBaselineSlowdownX}{0.31$\times$}

\newcommand{\SSNHLReasoningAcceptedPct}{25.1\%}

\newcommand{\SSNHLReasoningSlowdownX}{1.00$\times$}

\newcommand{\SSNHLReplaceAfterAcceptedPct}{33.7\%}

\newcommand{\SSNHLReplaceAfterSlowdownX}{0.96$\times$}

\newcommand{\SSNHLGlobalValidationAcceptedPct}{39.5\%}

\newcommand{\SSNHLGlobalValidationSlowdownX}{4.49$\times$}

\newcommand{\SSNHLLocalValidationAcceptedPct}{40.3\%}

\newcommand{\SSNHLLocalValidationSlowdownX}{3.15$\times$}

\newcommand{\SSNHLTrajectoryAcceptedPct}{83.5\%}

\newcommand{\SSNHLTrajectorySlowdownX}{2.78$\times$}

\newcommand{\ConductiveBaselineAcceptedPct}{1.6\%}

\newcommand{\ConductiveBaselineSlowdownX}{0.34$\times$}

\newcommand{\ConductiveReasoningAcceptedPct}{58.9\%}

\newcommand{\ConductiveReasoningSlowdownX}{1.00$\times$}

\newcommand{\ConductiveReplaceAfterAcceptedPct}{--}

\newcommand{\ConductiveReplaceAfterSlowdownX}{--}

\newcommand{\ConductiveGlobalValidationAcceptedPct}{--}

\newcommand{\ConductiveGlobalValidationSlowdownX}{--}

\newcommand{\ConductiveLocalValidationAcceptedPct}{--}

\newcommand{\ConductiveLocalValidationSlowdownX}{--}

\newcommand{\ConductiveTrajectoryAcceptedPct}{77.9\%}

\newcommand{\ConductiveTrajectorySlowdownX}{1.33$\times$}

\newcommand{\CombinedBaselineBalancedAccuracyPct}{28.1\%}

\newcommand{\CombinedReasoningBalancedAccuracyPct}{42.0\%}

\newcommand{\CombinedReplaceAfterBalancedAccuracyPct}{--}

\newcommand{\CombinedGlobalValidationBalancedAccuracyPct}{--}

\newcommand{\CombinedLocalValidationBalancedAccuracyPct}{--}

\newcommand{\CombinedTrajectoryBalancedAccuracyPct}{80.7\%}

\newcommand{\SSNHLReplaceAfterImprovedCount}{88}

\newcommand{\SSNHLReplaceAfterDegradedCount}{2}
\newcommand{\SSNHLReplaceAfterDeltaPP}{+8.6 pp}

\newcommand{\SSNHLGlobalValidationImprovedCount}{308}

\newcommand{\SSNHLGlobalValidationDegradedCount}{164}
\newcommand{\SSNHLGlobalValidationDeltaPP}{+14.4 pp}

\newcommand{\SSNHLLocalValidationImprovedCount}{272}

\newcommand{\SSNHLLocalValidationDegradedCount}{120}
\newcommand{\SSNHLLocalValidationDeltaPP}{+15.2 pp}

\newcommand{\SSNHLTrajectoryImprovedCount}{610}

\newcommand{\SSNHLTrajectoryDegradedCount}{26}
\newcommand{\SSNHLTrajectoryDeltaPP}{+58.4 pp}

\newcommand{\ConductiveTrajectoryImprovedPP}{+20.0 pp}

\newcommand{\ConductiveTrajectoryImprovedCount}{200}
\newcommand{\ConductiveTrajectoryDegradedPP}{+1.0 pp}

\newcommand{\ConductiveTrajectoryDegradedCount}{10}
\newcommand{\ConductiveTrajectoryDeltaPP}{+19.0 pp}

\newcommand{\RuntimeAvgInterventionsPerCase}{17.1}
\newcommand{\RuntimeAvoidInterventionsPerCase}{15.4}
\newcommand{\RuntimeAvgAlternativesTried}{3.70}
\newcommand{\RuntimeAlternativesForFiftyPctResolution}{2}
\newcommand{\RuntimeAlternativesForEightyPctResolution}{6}
\newcommand{\RuntimeRanOutOfAlternatives}{8.7\%}

\title{Answer Engineering: Local Trajectory Editing for Protocol-Constrained
Decision Making in Large Language Models}

\author{%
  Victor Lavrenko\\
  \small PeaceTech VC, Israel\\
  \small \texttt{victor@peacetech.vc}
  \and
  Anastasiia Molodnitskaia\\
  \small Rambam Health Care Campus, Israel\\
  \small \texttt{a\_molodnitskaia@rambam.health.gov.il}
}

\date{June 19, 2026}

\begin{document}

\maketitle

\begin{abstract}
  Large language models can produce confident but wrong answers when they
  violate procedures in domains where compliance is critical. Step-by-step
  reasoning improves structure and transparency but often shifts the
  distribution of errors rather than preventing them because procedures remain
  systematically underused. This effect arises across multiple protocol-driven
  domains; here, we evaluate it on a clinical benchmark requiring differential
  diagnosis using a binary adherence endpoint. In this task, compliant outcomes
  for the target condition decreased from 54.5\% under unguided generation to
  25.1\% under a step-by-step reasoning baseline, while performance on the
  contrastive condition increased from 1.6\% to 58.9\%. We hypothesize that
  reliability can be improved by intervening locally during generation,
  selecting valid reasoning trajectories and enforcing required tokens when
  violations occur. This hypothesis is implemented not as a new decoding
  algorithm but as a deterministic runtime and authoring layer over standard
  autoregressive generation, using rule-guided localized interventions to steer
  greedy decoding toward protocol-compliant trajectories without retraining or
  global search. In this workflow, localized trajectory intervention increased
  compliant outcomes for the target condition from 25.1\% to 83.5\% and
  increased protocol adherence on the contrastive condition from 58.9\% to
  77.9\%, suggesting that failures are consistent with structural violations
  that can be corrected through domain-specific protocols.
\end{abstract}

\section{Introduction}\label{sec:introduction}

Modern large language models (LLMs) generate fluent explanations and decisions,
and explicit reasoning has emerged as a widely used strategy for improving model
accuracy and reliability. Techniques such as chain-of-thought prompting,
structured reasoning, and agent-based planning have demonstrated measurable
performance gains across a range of tasks, particularly those requiring
multi-step inference or procedural decision-making
\parencite{wei2022cot,yao2023react,yao2023treeofthoughts,shinn2023reflexion}. As
a result, generating reasoning has become a common mechanism for improving both
model capability and interpretability.

However, subsequent research has shown that the presence of reasoning does not
guarantee that the reasoning is faithful to the decision process that produces
the final answer. Empirical studies have documented that language models
frequently generate explanations that function as rationalizations rather than
causal justifications of their outputs
\parencite{lyu2023faithfulcot,paul2024makingreasoningmatter}. In such cases,
reasoning steps may be internally inconsistent, incompatible with domain rules,
or disconnected from the mechanism that determines the final prediction. Even
models explicitly trained to produce structured reasoning remain only partially
faithful to the underlying decision process \parencite{yun2025medprm}. These
findings suggest that reasoning alone is not a sufficient reliability mechanism.
In practice, structured reasoning may shift the distribution of errors rather
than eliminate them, improving performance for some conditions while introducing
new procedural failures for others.

A practical manifestation of this limitation arises from the temporal structure
of generated reasoning. Models may commit to a decision early and then generate
reasoning steps that rationalize the chosen answer, producing a logically
inconsistent explanation of the decision process. Conversely, when reasoning is
generated before the decision, intermediate steps may be irrelevant, incomplete,
or disconnected from the mechanism that determines the final outcome. In both
cases, reasoning is present but does not reliably guide the decision process or
ensure protocol-compliant outcomes.

This limitation is particularly consequential in domains governed by explicit
procedural requirements. In regulated settings, correctness depends not only on
factual accuracy but also on adherence to mandated reasoning steps defined by
professional protocols. An answer may therefore appear plausible yet still be
operationally invalid if required checks or distinctions are omitted. Despite
their linguistic competence, current LLMs do not reliably enforce such protocol
constraints during generation
\parencite{wei2022cot,yao2023react,liu2023trustworthy,lyu2023faithfulcot}.

These observations suggest that improving reliability in protocol-driven domains
requires mechanisms that operate during generation rather than only after an
answer has been produced.

This work does not propose a new decoding algorithm. Instead, it presents a
deterministic runtime and authoring layer for localized trajectory intervention
during generation, preserving the underlying model and decoding process while
enabling protocol-guided control. Appendix~\ref{app:taxonomy} summarizes the
terminology and positioning used throughout the paper. 

The approach is positioned relative to prior work along three complementary
dimensions: failure coverage, intervention surface, and trajectory-time
capability. Unlike prompting, constrained decoding, or workflow-level
guardrails, Answer Engineering targets localized intervention within the
generated reasoning trajectory itself.

Evidence from clinical decision-making---one of the most extensively studied
protocol-governed domains---illustrates the magnitude of this challenge. In a
benchmark evaluating guideline adherence across 17 large language models and 20
diagnostic scenarios spanning 13 medical specialties, the best-performing model
achieved only 41.9/50 on a standardized protocol-adherence score
\parencite{fast2024amega}. In a separate study evaluating treatment-selection
decisions from real-world medical reports, agreement between model
recommendations and an expert heart team ranged from $\kappa = -0.47$ to $0.22$,
indicating poor concordance on realistic inputs
\parencite{roeschl2025guideline_concordant}. Additional studies report that LLMs
frequently fail to follow diagnostic or treatment guidelines in clinical
decision tasks and may generate confident recommendations that diverge from
established protocols \parencite{hager2024limitations_clinical_decision}.
Together these findings demonstrate that fluent reasoning does not guarantee
protocol-compliant outcomes.

Because LLM outputs are generated sequentially, reasoning unfolds as a
trajectory of intermediate statements that can serve as a runtime control
surface for enforcing procedural correctness. Let

\[
  \mathcal{R} = \{ r_1, \dots, r_n \}
\]

denote the reasoning trajectory generated by the model, where the tokens realizing each step are sampled from the same autoregressive
distribution $P_\theta$ conditioned on the
problem statement and the previously generated steps.

We use $\mathcal{R}$ for the human-readable sequence of reasoning steps and
$T=x_{1:t}$ for the token-level prefix operated on by the runtime.

In practice, reasoning trajectories may contain localized failures---for example,
when a required diagnostic distinction is omitted or a mandatory procedural step
is skipped. Such errors may affect both intermediate reasoning and the final
answer delivered to the user. Empirical evidence suggests that improving
reasoning faithfulness, verification, or structured evaluation often improves
downstream performance
\parencite{lyu2023faithfulcot,paul2024makingreasoningmatter,yun2025medprm}.
These observations suggest that some observed failures may reflect structural
violations of procedural requirements rather than insufficient domain knowledge
alone.

This observation motivates the need for mechanisms capable of intervening
directly within the reasoning trajectory during generation.

Such mechanisms must be configurable by domain experts and interpretable in
operational settings, where procedural requirements evolve and must remain
auditable. Accordingly, trajectory intervention is implemented through a small
set of explicit rule families that define permissible edits to the reasoning
trajectory during generation.

This work addresses these failures by applying localized runtime edits to the
reasoning trajectory during generation rather than after generation has been
completed.

To formalize this idea, we distinguish between the generated reasoning
trajectory and the controlled reasoning trajectory. Let

\[
  \mathcal{R}^* = \{ r_1^*, \dots, r_n^* \}
\]

denote the trajectory after runtime interventions. A controlled reasoning step
$r_i^*$ represents the accepted version of step $i$ after possible editing,
replacement, or regeneration to satisfy domain-specific constraints. The final
answer is produced from the controlled trajectory:

\[
  y \sim P_\theta(\cdot \mid s, \mathcal{R}^*)
\]

Local edits redefine the visible prefix used by the autoregressive model,
allowing generation to continue from a corrected state without requiring full
regeneration of the reasoning sequence. This mechanism operationalizes the
hypothesis that reliability can be improved by correcting localized protocol
violations during generation without retraining models or performing global
search.

A growing line of work has explored methods for improving reasoning and control
in LLMs, including prompting strategies and agent-based reasoning frameworks
\parencite{wei2022cot,yao2023react,yao2023treeofthoughts,shinn2023reflexion},
principle-based alignment and self-critique \parencite{bai2022constitutionalai},
and decoding-time control methods that guide generation behavior without
retraining, such as constrained or guided decoding and inference-time alignment
mechanisms
\parencite{yang2021fudge,shi2023contextawaredecoding,srivatsa2024deal}. These
approaches primarily influence generation globally or probabilistically. In
contrast, the method proposed in this paper intervenes locally during generation
by enforcing step-level protocol rules within the reasoning trajectory itself.

A critical open question is whether enforcing procedural constraints during
generation can measurably improve protocol-compliant outcomes compared with
reasoning alone.

To evaluate this question, we apply the approach to a controlled clinical
reasoning benchmark involving sudden sensorineural hearing loss (SSNHL), where
protocol violations are clinically consequential and operationally measurable.
The benchmark uses template-generated paired cases evaluated under fixed decoding
conditions and compares
unguided generation, reasoning-prompt generation, and trajectory-controlled
generation to measure the effect of runtime protocol enforcement under matched
conditions.

Although the empirical evaluation in this paper focuses on a clinical protocol,
the trajectory-control mechanism itself is not domain-specific. The framework is
implemented as an executable runtime system with observable telemetry and
rule-based configuration. For example, the repository quickstart notebook
includes a software-engineering prompt about rate limiting in a small Python web
API service, where a local replacement rule redirects a from-scratch
implementation plan toward reuse of an existing library.\footnote{See the
repository quickstart notebook \path{notebooks/quickstart.ipynb}.}

This work therefore investigates whether localized trajectory intervention can
provide a runtime control surface for improving protocol-compliant reasoning.

To address these requirements, this paper makes three contributions:

\begin{enumerate}[leftmargin=*,itemsep=0.2em]

  \item A decoding-time runtime mechanism for enforcing protocol compliance
    through localized trajectory intervention during generation, without
    retraining models or performing global search.

  \item A rule-based control abstraction for specifying domain-specific
    procedural constraints as interpretable trajectory-editing operations.

  \item An empirical evaluation demonstrating that step-by-step reasoning alone
    may reduce protocol compliance in structured decision tasks and that
    localized trajectory intervention substantially improves protocol-compliant
    outcomes.

\end{enumerate}

\section{Failure Modes in Protocol-Constrained
Generation}\label{sec:failure_modes}

Protocol-constrained reasoning requires not only factual correctness but also
explicit adherence to externally defined procedural steps. Empirical studies
suggest that language models often possess relevant domain knowledge yet fail to
consistently apply required procedural checks during generation. Analyses of
reasoning faithfulness and guideline adherence indicate that such violations
frequently arise within the visible reasoning trajectory itself---for example
when intermediate steps omit required distinctions or skip mandatory
verification procedures
\parencite{lanham2023measuringfaithfulnesscot,arcuschin2025cotwild,
fast2024amega}. Despite strong benchmark performance, LLMs therefore often
produce fluent answers that violate protocol requirements. We summarize five
recurring failure modes (F1--F5) documented in the literature.

\subsection*{F1: Unfaithful Visible
Reasoning}\label{sec:f1_unfaithful_reasoning}

Reasoning traces produced by language models are often not faithful to the
decision process that generates the final answer. Studies of chain-of-thought
reasoning show that models can produce plausible intermediate explanations that
do not reflect the actual factors driving the prediction
\parencite{lanham2023measuringfaithfulnesscot,lyu2023faithfulcot}. Subsequent
work shows that explanations generated in the wild frequently diverge from the
model's internal computation and may function primarily as post-hoc
rationalizations \parencite{arcuschin2025cotwild}.

\subsection*{F2: Prompt and Stem Drift}\label{sec:f2_prompt_stem_drift}

During multi-step generation, the reasoning trajectory may gradually diverge
from the original task specification. As generation progresses, the model may
omit critical constraints, forget key details from the case description, or
reinterpret variables in ways that detach the reasoning from the original
prompt. Such drift can arise in multi-step reasoning frameworks, including
tree-based and agent-based approaches, because long reasoning trajectories may
gradually detach from the original task specification
\parencite{yao2023treeofthoughts,yao2023react}.

\subsection*{F3: Rationalization of Chosen
Answers}\label{sec:f3_rationalization}

Language models may generate explanations that justify an answer rather than
derive it. In such cases, the explanation is constructed to rationalize a
prediction that has already been selected, rather than reflecting the reasoning
process that produced it. This behavior is related to unfaithful reasoning
traces but differs in that the explanation serves primarily as post-hoc
justification. In protocol-constrained settings, this can be especially
problematic because the resulting explanation may appear systematic while
failing to actually perform the required procedural analysis
\parencite{lanham2023measuringfaithfulnesscot,arcuschin2025cotwild}.

\subsection*{F4: Local Protocol
Inconsistencies}\label{sec:f4_local_protocol_inconsistencies}

A recurring failure pattern in LLM reasoning is that an explanation may remain
globally plausible while containing a local error at a protocol-critical step.
Even when the reasoning trajectory broadly follows the intended analysis, a
single incorrect inference or protocol misinterpretation can invalidate the
resulting conclusion. In protocol-constrained tasks, such errors may involve
misstating a key distinction, using terminology that does not match the
triggering condition, or introducing a locally inconsistent inference
\parencite{turpin2023language,lyu2023faithfulcot,
paul2024makingreasoningmatter,wiegreffe2024faithfulness}.

\subsection*{F5: Omission of Mandatory Protocol
Steps}\label{sec:f5_missing_protocol_steps}

Language models can also omit procedural steps required by domain protocols.
Even when a model correctly interprets much of the problem, its generated
reasoning may fail to explicitly perform a required verification step,
diagnostic distinction, or safety check. This pattern is consistent with the
broader observation that many current reasoning methods improve answer quality
without explicitly guaranteeing satisfaction of externally specified procedural
constraints \parencite{yao2023react,yao2023treeofthoughts,srivatsa2024deal,
bai2022constitutionalai}.

Taken together, these failure modes suggest that protocol violations frequently
arise within the visible reasoning trajectory itself. Models may drift from the
case description, rationalize a chosen conclusion, introduce a local
inconsistency, or omit a required procedural step. This motivates mechanisms
that intervene during generation rather than relying solely on prompting or
post-hoc validation.

\section{Existing Approaches and Challenges for Protocol\hyp{}Constrained
Generation}

A large body of work has improved the reliability of large language models
(LLMs) through prompting strategies, structured reasoning frameworks, sampling
methods, constrained decoding, and symbolic control
\parencite{wei2022cot,yao2023react,wang2023selfconsistency,
zhou2025structured,liu2023trustworthy,zhao2024ctg}. These approaches achieve
substantial gains on reasoning benchmarks and remain strong baselines for many
applications.

However, when evaluated in protocol-governed settings, these methods address
protocol compliance only indirectly. Protocol-constrained reasoning requires not
only producing a correct final answer but also performing specific procedural
checks or reasoning steps required by external rules, guidelines, or
regulations.

When analyzed through the lens of the failure modes introduced in
Section~\ref{sec:failure_modes}---unfaithful reasoning (F1), prompt drift (F2),
rationalization (F3), local protocol inconsistencies (F4), and omission of
mandatory protocol steps (F5)---existing approaches exhibit uneven coverage.
Table~\ref{tab:failure-mode-comparison} summarizes how representative classes of
methods mitigate these failure modes.

\begin{table*}[!htbp]
  \centering
  \small
  \begin{tblr}{ colspec = {Q[l,10.5cm]ccccc}, row{1} = {font=\bfseries}, hlines
    } Method class & F1 & F2 & F3 & F4 & F5 \\

    Prompting / CoT \parencite{wei2022cot} & \partialcov & \partialcov &
    \partialcov & \none & \none \\

    Self-consistency / sampling \parencite{wang2023selfconsistency} &
    \partialcov & \partialcov & \partialcov & \none & \none \\

    ReAct / agent reasoning \parencite{yao2023react,shinn2023reflexion} &
    \partialcov & \partialcov & \partialcov & \none & \none \\

    Structured reasoning / scaffolds
    \parencite{zhou2025structured,yao2023treeofthoughts} & \partialcov &
    \partialcov & \partialcov & \none & \none \\

    Verification / correction / calibration
    \parencite{deng2023answercalibration,tyen2024cannotfind} & \partialcov &
    \none & \partialcov & \none & \none \\

    Principle-based alignment / self-critique
    \parencite{bai2022constitutionalai} & \partialcov & \partialcov &
    \partialcov & \none & \none \\

    Process reward models / trajectory scoring
    \parencite{lightman2023prm800k,yun2025medprm} & \partialcov & \partialcov &
    \partialcov & \partialcov & \none \\

    Structured-output and guided decoding
    \parencite{yang2021fudge,shi2023contextawaredecoding,koo2024automata,zhao2024ctg,
    srivatsa2024deal} & \none & \none & \none & \partialcov & \partialcov \\

    Rule-based / expert systems \parencite{liu2023trustworthy,zhao2024ctg} &
    \none & \none & \none & \full & \full \\

    \midrule
    \textbf{Answer Engineering (this work)}
    & \partialcov & \full & \partialcov & \full & \partialcov \\

  \end{tblr}

  \caption{Coverage of protocol failure modes by representative classes of
    approaches. F1: unfaithful reasoning; F2: prompt drift; F3: rationalization;
    F4: local protocol inconsistencies; F5: omission of mandatory protocol
    steps. Legend: \full~strong coverage, \partialcov~partial or indirect
    mitigation, \none~no direct coverage. For constrained decoding, partial
    coverage applies only when mandatory protocol elements can be encoded
    explicitly in the constrained output
    representation.}\label{tab:failure-mode-comparison}
\end{table*}

Beyond empirical coverage of failure modes, existing approaches can also be
understood according to the \emph{mechanism by which they influence the
reasoning trajectory}. Broadly, three intervention paradigms appear in the
literature:

\begin{itemize}[leftmargin=*]
  \item \textbf{Generation biasing}, which influences reasoning indirectly
    through prompts, instructions, or structural scaffolds that encourage
    certain reasoning patterns.

  \item \textbf{Trajectory selection}, which generates multiple candidate
    reasoning paths and selects or reranks them according to sampling criteria
    or learned reward models.

  \item \textbf{Explicit constraint enforcement}, which imposes structural or
    symbolic rules on the reasoning process itself.
\end{itemize}

These paradigms correspond to different control points in the generation
pipeline and therefore differ in how directly they can enforce explicitly
specified protocol constraints.

While Table~\ref{tab:failure-mode-comparison} summarizes empirical coverage of
protocol-related failure modes, these approaches intervene at different stages
of the reasoning process. Some bias generation through prompts or sampling
strategies, others evaluate or rerank candidate trajectories after they are
produced, and symbolic systems replace learned reasoning with explicit decision
rules.

Table~\ref{tab:trajectory_operations} therefore reorganizes several of the above
method families according to their \emph{mechanistic intervention point}. Rather
than reproducing the full taxonomy of Table~\ref{tab:failure-mode-comparison},
the table highlights representative intervention styles that differ most clearly
in how they influence the generation process.

\begin{table*}[t]
  \centering
  \scriptsize
  \setlength{\tabcolsep}{3.0pt}
  \renewcommand{\arraystretch}{1.12}

  \caption{Trajectory-time reasoning operations in LLM-based generation
  methods.}\label{tab:trajectory_operations}

  \begin{tabularx}{\textwidth}{ >{\raggedright\arraybackslash}p{0.23\textwidth}
      >{\raggedright\arraybackslash}p{0.23\textwidth} ccccc
      >{\raggedright\arraybackslash}X }
    \toprule

    Method & Primary mechanism & \multicolumn{5}{c}{Ops.$^{\dagger}$} & Main
    limitation \\

    \cmidrule(lr){3-7}

    & &
    \textbf{1}
    & \textbf{2} & \textbf{3} & \textbf{4} & \textbf{5} & \\

    \midrule

    Prompting / Chain-of-Thought & Prompt instructions and demonstrations &
    \cmark & \xmark & \xmark & \xmark & \xmark & Encourages structured reasoning
    but cannot enforce or repair incorrect steps
    \parencite{wei2022cot,wang2023selfconsistency} \\

    Reflection / retry & Iterative self-critique after answer generation &
    \partialmark & \partialmark & \partialmark & \partialmark & \partialmark &
    Repairs occur after an answer exists rather than during generation
    \parencite{shinn2023reflexion,bai2022constitutionalai} \\

    Verification / auditing & External validators or checkers & \xmark &
    \partialmark & \xmark & \partialmark & \xmark & Detects violations but
    typically does not alter reasoning in real time
    \parencite{lanham2023measuringfaithfulnesscot,lyu2023faithfulcot} \\

    Constrained decoding & Token-level constraints or classifiers & \partialmark
    & \partialmark & \partialmark & \xmark & \cmark & Controls token
    admissibility rather than reasoning validity
    \parencite{yang2021fudge,koo2024automata,srivatsa2024deal} \\

    Trajectory search & Selection among sampled reasoning paths & \partialmark &
    \partialmark & \xmark & \xmark & \partialmark & Chooses among alternatives
    but does not locally repair a trajectory
    \parencite{yao2023treeofthoughts,wang2023selfconsistency} \\

    \textbf{Answer Engineering}
    & \textbf{Local protocol constraints applied during decoding} &
    \textbf{\cmark} & \textbf{\cmark} & \textbf{\cmark} & \textbf{\cmark} &
    \textbf{\cmark} & Requires authored protocol rules but preserves
    probabilistic model flexibility \\

    \bottomrule
  \end{tabularx}

  \vspace{0.6em}

  % -------- Operations definitions --------

  \begin{tabularx}{\textwidth}{ >{\raggedright\arraybackslash}p{0.035\textwidth}
    >{\raggedright\arraybackslash}X }
    \toprule
    \multicolumn{2}{l}{\textbf{Trajectory-time operations$^{\dagger}$}} \\
    \midrule

    \textbf{1} & Require explicit step-by-step reasoning during generation. \\

    \textbf{2} & Avoid reasoning steps that violate protocol or domain
    constraints. \\

    \textbf{3} & Modify or replace the reasoning step currently being generated.
    \\

    \textbf{4} & Retrospectively modify previously generated reasoning before
    generation continues. \\

    \textbf{5} & Resume generation from the corrected reasoning trajectory. \\

    \bottomrule
  \end{tabularx}

  \vspace{0.25em}

  {\footnotesize \cmark{} = direct support;\quad \partialmark{} = partial
  support;\quad \xmark{} = not supported. }

  \vspace{0.25em}

  {\footnotesize $^{\dagger}$Trajectory-time operations act directly on the
    evolving reasoning trajectory during generation, rather than only on
    prompts, outputs, or candidate selection. }

\end{table*}

\begin{table*}[t]
\centering
\scriptsize
\setlength{\tabcolsep}{2.8pt}
\renewcommand{\arraystretch}{1.24}
\caption{Control surfaces for protocol-constrained generation. Methods differ in
intervention timing, control surface, authoring model, and the guarantees they
provide.}\label{tab:control-surfaces}
\begin{tabularx}{\textwidth}{
>{\raggedright\arraybackslash}p{0.15\textwidth}
>{\raggedright\arraybackslash}p{0.19\textwidth}
>{\raggedright\arraybackslash}p{0.11\textwidth}
>{\raggedright\arraybackslash}p{0.12\textwidth}
>{\raggedright\arraybackslash}p{0.10\textwidth}
>{\raggedright\arraybackslash}p{0.13\textwidth}
>{\raggedright\arraybackslash}X}
\toprule
Control surface &
Representative systems &
Intervention timing &
Modified object &
Repairs generated text &
Authoring model &
Primary control property \\
\midrule
Prompt / reasoning scaffolds &
CoT, ReAct, Tree-of-Thoughts &
before generation &
prompt or scaffold &
No &
prompts / examples &
reasoning guidance \\
\addlinespace
Structured / constrained decoding &
Guidance, LMQL, Outlines, grammar/schema decoding, SGLang &
during decoding &
token admissibility or decode program &
No &
grammar / schema / program &
structural validity \\
\addlinespace
Post-hoc validation / correction &
Self-Refine, Reflexion, validation pipelines &
after generation &
completed output &
Yes (post-hoc) &
validators / retry workflows &
output filtering or correction \\
\addlinespace
Runtime guardrails / orchestration &
NeMo Guardrails, workflow policies &
runtime / workflow boundary &
interaction policy &
limited &
policy / workflow rules &
workflow safety \\
\addlinespace
\textbf{Trajectory intervention (this work)} &
\textbf{Answer Engineering} &
\textbf{during generation} &
\textbf{generated reasoning trajectory} &
\textbf{Yes (local)} &
\textbf{local protocol rules} &
\textbf{local protocol-constraint enforcement} \\
\bottomrule
\end{tabularx}
\end{table*}

\subsection{Relation to Existing Control Surfaces}

Recent runtime frameworks expose multiple control surfaces for influencing
language-model behavior, including prompt scaffolds, structured decoding
constraints, post-hoc validation, and workflow-level guardrails.
Table~\ref{tab:control-surfaces} summarizes these intervention surfaces and
situates Answer Engineering within this broader runtime-control landscape.
Unlike prompt- or workflow-level methods, Answer Engineering targets localized
intervention within the evolving reasoning trajectory during generation.

\paragraph{Structured-output and guided-decoding systems.}
Modern inference engines such as vLLM and SGLang support structured-output
generation using grammars, JSON schemas, regular expressions, and other
admissibility constraints \parencite{kwon2023vllm,zheng2024sglang}. These
mechanisms ensure that generated text satisfies formal structural requirements
and are widely used in production systems for enforcing output validity.
Libraries such as Guidance, LMQL, Outlines, and related
constrained-decoding toolkits provide programmable interfaces for structured
generation that allow developers to define grammar-based, schema-based, or
programmatic constraints on the generation process
\parencite{guidance2023,beurer2023lmql,koo2024automata}.

These systems are highly effective when correctness can be expressed as
structural admissibility of token sequences or outputs. In contrast, Answer
Engineering targets semantic violations arising within otherwise structurally
valid reasoning trajectories.

\paragraph{Programmable guardrail and orchestration systems.}
Frameworks such as NeMo Guardrails provide configurable policies governing model
inputs, outputs, and tool interactions \parencite{nemoguardrails2023}. These
systems enable developers to define workflow-level rules, safety checks, and
intervention logic that can redirect or block model behavior based on
application policies. Such guardrail mechanisms are widely used to improve
system reliability and enforce domain-specific constraints in deployed
applications.

Guardrail systems typically intervene at workflow boundaries by filtering,
redirecting, or blocking interactions. Answer Engineering instead modifies the
reasoning trajectory itself and resumes generation from the corrected prefix.

\paragraph{Positioning within the runtime control landscape.}
These methods intervene at different stages of generation and therefore address
different classes of constraints and failure modes. Structured-output and
guided-decoding systems enforce formal admissibility, guardrail frameworks
regulate workflow behavior, and trajectory intervention modifies the evolving
reasoning process itself.

Prompt-based methods such as chain-of-thought prompting influence generation by
encouraging models to produce intermediate explanations that structure the
reasoning process \parencite{wei2022cot}. Sampling strategies such as
self-consistency further improve robustness by selecting answers that remain
stable across multiple reasoning trajectories
\parencite{wang2023selfconsistency}. Agent-style reasoning frameworks such as
ReAct combine reasoning with tool use and external actions, often improving
task-solving performance \parencite{yao2023react}. Structured reasoning
frameworks attempt to organize reasoning into more explicit intermediate
structures \parencite{zhou2025structured}. While these approaches can reduce
some forms of reasoning instability, they do not by themselves guarantee
protocol compliance during generation.

Process reward models and trajectory\hyp{}scoring methods evaluate intermediate
reasoning steps using le\-arned reward functions and external evidence,
preferring trajectories that receive higher scores
\parencite{lightman2023prm800k,yun2025medprm}. These methods can improve overall
reasoning quality and alignment with domain knowledge, but they typically
operate by selecting among candidate trajectories rather than directly repairing
violations within the evolving reasoning process.

Structured-output and guided decoding methods enforce structural constraints on
model outputs, for example ensuring that generated text satisfies a grammar or
schema \parencite{koo2024automata}. These approaches are especially effective
when correctness depends on formal structure, such as grammars or schemas. Their
applicability to context-sensitive protocol constraints depends on whether those
constraints can be encoded explicitly enough at decoding time.

Expert systems and rule-based controllers enforce domain protocols through
explicit symbolic rule structures \parencite{liu2023trustworthy,zhao2024ctg}.
These systems can guarantee the presence of mandatory protocol steps, but they
can require substantial symbolic specification and maintenance effort as
protocols evolve.

These comparisons position existing methods along three distinct dimensions:
failure coverage, intervention surface, and trajectory-time capability.
Prompting and sampling influence reasoning indirectly, constrained decoding
enforces structural admissibility, guardrail systems regulate workflow behavior,
and Answer Engineering intervenes locally within the generated reasoning
trajectory itself.

These limitations motivate the approach summarized in
Tables~\ref{tab:failure-mode-comparison},~\ref{tab:control-surfaces},
and~\ref{tab:trajectory_operations}. In
this work we develop \emph{Answer Engineering}, a decoding-time framework that
enforces protocol constraints through local intervention in the evolving
reasoning trajectory using declarative domain rules. The trajectory-control
mechanism is described in the following sections.

\section{Decoding-Time Control of Reasoning
Trajectories}\label{sec:trajectory_control}

Section~\ref{sec:introduction} introduced trajectory control as the general
decoding-time principle and trajectory editing as its concrete runtime mechanism.
This section formalizes the approach and explains how prefix-conditioned decoding
enables localized interventions during generation.

\tikzset{ ae/token/.style={ draw, rounded corners=1.5pt, minimum height=5.5mm,
  minimum width=8mm, inner sep=1.5pt }, ae/trigger/.style={ ae/token,
  fill=black!10 }, ae/scope/.style={ ae/token, fill=black!6 },
  ae/chosen/.style={ ae/token, fill=black!18 }, ae/note/.style={
  font=\footnotesize, align=center }, ae/tinylabel/.style={ font=\scriptsize,
  align=center }, ae/mainarrow/.style={ ->, thick } }

\subsection{Prefix-conditioned decoding}\label{sec:prefix-conditioned}

Answer Engineering relies on a basic property of autoregressive decoding:
future-token distributions depend on the visible prefix. Therefore, local edits
to the generated trajectory redefine the context for subsequent generation
without retraining the model. This makes runtime trajectory correction feasible
for protocol-constrained reasoning.\footnote{Implementation detail: after an
edit, the KV cache is rebuilt from the earliest modified token. Because the
cache is only an efficiency mechanism, rebuilding from the edited prefix yields
the same continuation distribution as fresh decoding from that prefix. See
Appendix~\ref{app:implementation} for a brief implementation note.}

\paragraph{Observation: Prefix-conditioned continuation.}
In an autoregressive transformer, the probability of future tokens depends only
on the visible prefix. Let $P_\theta(x_{t+1}\mid x_{1:t})$ denote the next-token
distribution of an autoregressive language model. Then the distribution of all
future tokens $x_{t+1:T}$ depends only on the visible prefix $x_{1:t}$.
In particular,
if a subsequence $x_{i:j}$ in the prefix is replaced by an edited segment
$\tilde{x}_{i:j}$, the future generation distribution becomes
\[
  P_\theta(x_{t+1:T}\mid x_{1:i-1},\tilde{x}_{i:j},x_{j+1:t}),
\]
which is identical to the distribution obtained if the edited prefix had been
generated directly by the model.

This autoregressive factorization implies that editing the visible prefix
changes the context for all subsequent decoding without retraining. Accordingly,
Answer Engineering can enforce protocol knowledge through local trajectory edits
(Figures~\ref{fig:ae_retroactive_edit}--\ref{fig:ae_force_future}) and primarily
targets the failure modes F2--F5.

\subsection{The Trajectory Control
Principle}\label{sec:trajectory_control_principle}
\textbf{Addresses: F2, F3, F4, F5}

Because generation is conditioned only on the visible prefix, a runtime
controller can intervene at any point in the reasoning trajectory. These
interventions operate directly on the token sequence and can modify previously
generated tokens, redirect local continuations, or enforce specific future
tokens.

\begin{figure*}[!htbp]
  \centering
  \resizebox{0.8\textwidth}{!}{%
    \begin{tikzpicture}[font=\small, >=Latex]

      % =========================================================
      % Stage 1: generated trajectory
      % =========================================================
      \node[ae/note] at (1.5,0.75) {already generated trajectory};

      \node[ae/token]   (x1) at (0.0,0) {$x_1$}; \node[ae/token]   (x2) at
      (1.1,0) {$x_2$}; \node[ae/trigger] (x3) at (2.2,0) {$x_3$};
      \node[ae/scope]   (x4) at (3.3,0) {$x_4$}; \node[ae/scope]   (x5) at
      (4.4,0) {$x_5$}; \node[ae/trigger] (x6) at (5.5,0) {$x_6$};
      \node[ae/token]   (x7) at (6.6,0) {$x_7$}; \node[ae/trigger] (x8) at
      (7.7,0) {$x_8$}; \node[ae/token]   (dots1) at (8.8,0) {$\cdots$};

      \draw[decorate,decoration={brace,amplitude=4pt}] (1.8,0.42) -- (8.1,0.42)
      node[midway,above=6pt,ae/tinylabel] {guard scope};

      \draw[densely dashed, rounded corners=2pt] (2.85,-0.38) rectangle
      (4.85,0.38); \node[ae/tinylabel] at (3.85,-0.7) {edit span};

      \node[ae/tinylabel] at (2.2,-0.5) {trigger}; \node[ae/tinylabel] at
      (5.5,-0.5) {trigger}; \node[ae/tinylabel] at (7.7,-0.5) {trigger};

      \draw[ae/mainarrow] (3.85,-1.1) -- (3.85,-2.0);

      % =========================================================
      % Stage 2: candidate proposals
      % =========================================================
      \node[ae/note] at (3.85,-2.5) {manual candidate replacements};

      % candidate 1
      \node[ae/token] (c11) at (1.5,-3.25) {$c^{(1)}_1$}; \node[ae/token] (c12)
      at (2.7,-3.25) {$c^{(1)}_2$}; \node[ae/token] (c13) at (3.9,-3.25)
      {$c^{(1)}_3$}; \node[ae/token] (c14) at (5.1,-3.25) {$c^{(1)}_4$};
      \node[ae/tinylabel, anchor=west] at (6.1,-3.25) {score $s_1$};

      % candidate 2
      \node[ae/token] (c21) at (1.5,-4.15) {$c^{(2)}_1$}; \node[ae/token] (c22)
      at (2.7,-4.15) {$c^{(2)}_2$}; \node[ae/tinylabel, anchor=west] at
      (6.1,-4.15) {score $s_2$};

      % best candidate
      \node[ae/chosen] (c31) at (1.5,-5.05) {$c^{(*)}_1$}; \node[ae/chosen]
      (c32) at (2.7,-5.05) {$c^{(*)}_2$}; \node[ae/chosen] (c33) at (3.9,-5.05)
      {$c^{(*)}_3$}; \node[ae/tinylabel, anchor=west] at (6.1,-5.05) {highest
      score};

      % scoring note
      \node[ae/note] at (10.0,-4.15) {candidates scored\\by model likelihood};

      \draw[ae/mainarrow] (3.85,-5.6) -- (3.85,-6.5);

      % =========================================================
      % Stage 3: edited trajectory
      % =========================================================
      \node[ae/note] at (3.85,-6.95) {best candidate inserted};

      \node[ae/token]  (y1) at (0.0,-7.7) {$x_1$}; \node[ae/token]  (y2) at
      (1.1,-7.7) {$x_2$}; \node[ae/token]  (y3) at (2.2,-7.7) {$x_3$};

      \node[ae/chosen] (y41) at (3.3,-7.7) {$c^{(*)}_1$}; \node[ae/chosen] (y42)
      at (4.5,-7.7) {$c^{(*)}_2$}; \node[ae/chosen] (y43) at (5.7,-7.7)
      {$c^{(*)}_3$};

      \node[ae/token]  (y5) at (6.8,-7.7) {$x_6$}; \node[ae/token]  (y6) at
      (7.9,-7.7) {$x_7$}; \node[ae/token]  (y7) at (9.0,-7.7) {$x_8$};
      \node[ae/token]  (dots2) at (10.1,-7.7) {$\cdots$};

      \draw[ae/mainarrow] (3.3,-8.15) -- (3.3,-9.05); \node[ae/note] at
      (3.3,-9.6) {rebuild KV cache\\from earliest edited token};

      \draw[ae/mainarrow] (10.65,-7.7) -- (11.8,-7.7); \node[ae/note,
      anchor=west] at (11.9,-7.7) {resume decoding};

    \end{tikzpicture}%
  } \caption{\textbf{Retroactive span editing.} A triggered local span is
    replaced with the highest-scoring protocol-valid candidate, then decoding
    resumes from the rebuilt prefix.}\label{fig:ae_retroactive_edit}
\end{figure*}
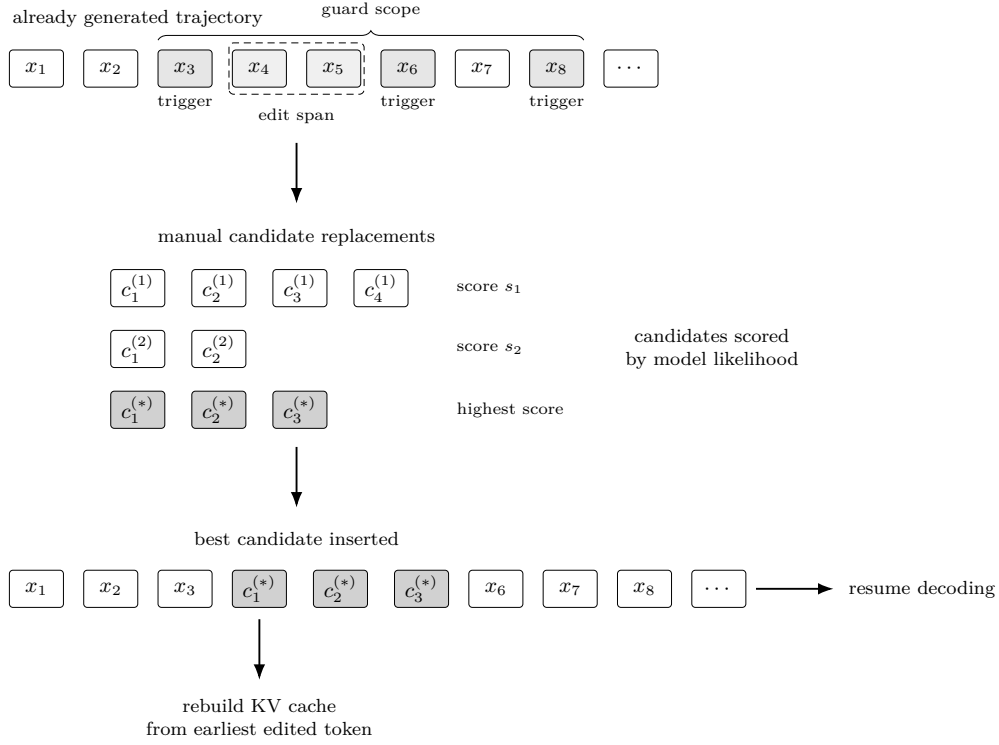

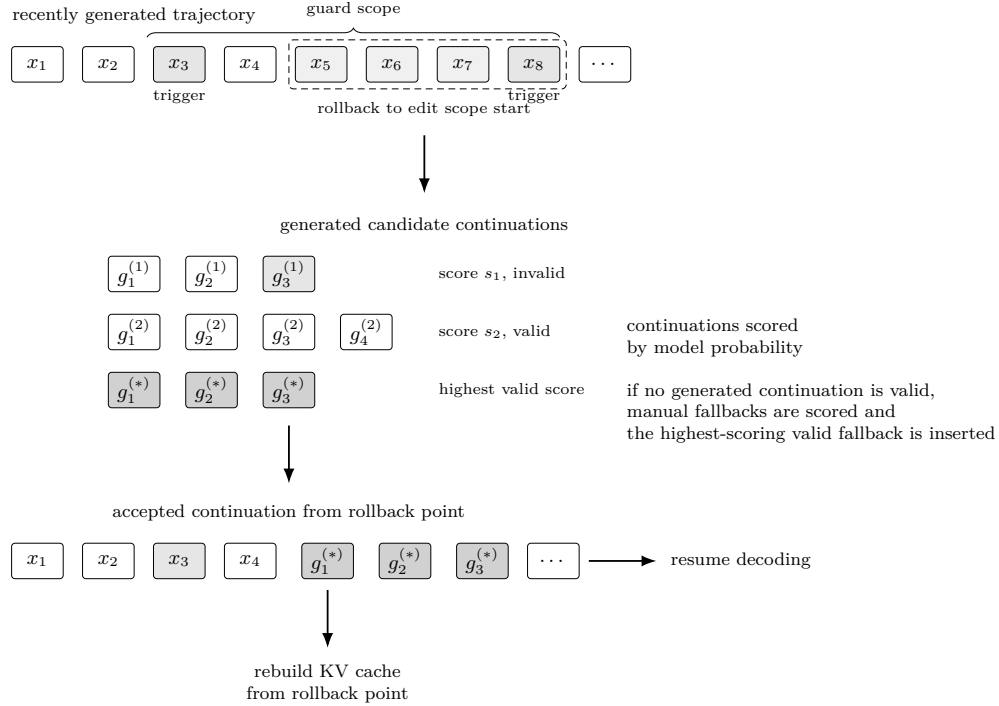
\begin{figure*}[!htbp]
  \centering
  \resizebox{0.8\textwidth}{!}{%
    \begin{tikzpicture}[font=\small, >=Latex]

      % =========================================================
      % Stage 1: generated trajectory
      % =========================================================
      \node[ae/note] at (1.5,0.75) {recently generated trajectory};

      \node[ae/token]   (x1) at (0.0,0) {$x_1$}; \node[ae/token]   (x2) at
      (1.1,0) {$x_2$}; \node[ae/trigger] (x3) at (2.2,0) {$x_3$};
      \node[ae/token]   (x4) at (3.3,0) {$x_4$}; \node[ae/scope]   (x5) at
      (4.4,0) {$x_5$}; \node[ae/scope]   (x6) at (5.5,0) {$x_6$};
      \node[ae/scope]   (x7) at (6.6,0) {$x_7$}; \node[ae/trigger] (x8) at
      (7.7,0) {$x_8$}; \node[ae/token]   (dots1) at (8.8,0) {$\cdots$};

      \draw[decorate,decoration={brace,amplitude=4pt}] (1.7,0.42) -- (8.1,0.42)
      node[midway,above=6pt,ae/tinylabel] {guard scope};

      \draw[densely dashed, rounded corners=2pt] (3.9,-0.38) rectangle
      (8.2,0.38); \node[ae/tinylabel] at (6.0,-0.7) {rollback to edit scope
      start};

      \node[ae/tinylabel] at (2.2,-0.5) {trigger}; \node[ae/tinylabel] at
      (7.7,-0.5) {trigger};

      \draw[ae/mainarrow] (6.0,-1.1) -- (6.0,-2.0);

      % =========================================================
      % Stage 2: generated continuations
      % =========================================================
      \node[ae/note] at (6.0,-2.5) {generated candidate continuations};

      % invalid candidate 1
      \node[ae/token]   (g11) at (1.5,-3.25) {$g^{(1)}_1$}; \node[ae/token]
      (g12) at (2.7,-3.25) {$g^{(1)}_2$}; \node[ae/trigger] (g13) at (3.9,-3.25)
      {$g^{(1)}_3$}; \node[ae/tinylabel, anchor=west] at (6.1,-3.25) {score
      $s_1$, invalid};

      % valid candidate 2
      \node[ae/token]   (g21) at (1.5,-4.15) {$g^{(2)}_1$}; \node[ae/token]
      (g22) at (2.7,-4.15) {$g^{(2)}_2$}; \node[ae/token]   (g23) at (3.9,-4.15)
      {$g^{(2)}_3$}; \node[ae/token]   (g24) at (5.1,-4.15) {$g^{(2)}_4$};
      \node[ae/tinylabel, anchor=west] at (6.1,-4.15) {score $s_2$, valid};

      % chosen generated candidate
      \node[ae/chosen]  (g31) at (1.5,-5.05) {$g^{(*)}_1$}; \node[ae/chosen]
      (g32) at (2.7,-5.05) {$g^{(*)}_2$}; \node[ae/chosen]  (g33) at (3.9,-5.05)
      {$g^{(*)}_3$}; \node[ae/tinylabel, anchor=west] at (6.1,-5.05) {highest
      valid score};

      % note
      \node[ae/note, align=left] at (12.0,-4.9) {continuations scored\\by model
        probability\\\\if no generated continuation is valid,\\manual fallbacks
        are scored and\\the highest-scoring valid fallback is inserted};

      \draw[ae/mainarrow] (3.9,-5.6) -- (3.9,-6.5);

      % =========================================================
      % Stage 3: accepted continuation / fallback path
      % =========================================================
      \node[ae/note] at (3.9,-6.95) {accepted continuation from rollback point};

      \node[ae/token]   (y1) at (0.0,-7.7) {$x_1$}; \node[ae/token]   (y2) at
      (1.1,-7.7) {$x_2$}; \node[ae/trigger] (y3) at (2.2,-7.7) {$x_3$};
      \node[ae/token]   (y4) at (3.3,-7.7) {$x_4$};

      \node[ae/chosen]  (y51) at (4.5,-7.7) {$g^{(*)}_1$}; \node[ae/chosen]
      (y52) at (5.7,-7.7) {$g^{(*)}_2$}; \node[ae/chosen]  (y53) at (6.9,-7.7)
      {$g^{(*)}_3$};

      \node[ae/token]   (dots2) at (8.0,-7.7) {$\cdots$};

      \draw[ae/mainarrow] (4.5,-8.15) -- (4.5,-9.05); \node[ae/note] at
      (4.5,-9.6) {rebuild KV cache\\from rollback point};

      \draw[ae/mainarrow] (8.55,-7.7) -- (9.6,-7.7); \node[ae/note, anchor=west]
      at (9.7,-7.7) {resume decoding};

    \end{tikzpicture}%
  } \caption{\textbf{Local rollback and continuation probing.} Decoding rolls
    back to a local scope, evaluates candidate continuations, accepts the
    highest-scoring valid one (or a fallback), and resumes from the repaired
    prefix.}\label{fig:ae_local_rollback}
\end{figure*}

\begin{figure*}[!htbp]
  \centering
  \resizebox{0.8\textwidth}{!}{%
    \begin{tikzpicture}[font=\small, >=Latex]

      % =========================================================
      % Stage 1: generated trajectory
      % =========================================================
      \node[ae/note] at (1.5,0.75) {already generated trajectory};

      \node[ae/token]   (x1) at (0.0,0) {$x_1$}; \node[ae/token]   (x2) at
      (1.1,0) {$x_2$}; \node[ae/trigger] (x3) at (2.2,0) {$x_3$};
      \node[ae/token]   (x4) at (3.3,0) {$x_4$}; \node[ae/token]   (x5) at
      (4.4,0) {$x_5$}; \node[ae/token]   (x6) at (5.5,0) {$x_6$};
      \node[ae/trigger] (x7) at (6.6,0) {$x_7$}; \node[ae/token]   (dots1) at
      (7.7,0) {$\cdots$};

      \draw[decorate,decoration={brace,amplitude=4pt}] (1.8,0.42) -- (7.0,0.42)
      node[midway,above=6pt,ae/tinylabel] {guard scope};

      \node[ae/tinylabel] at (2.2,-0.5) {trigger}; \node[ae/tinylabel] at
      (6.6,-0.5) {trigger};

      \node[ae/tinylabel] at (7.7,-0.9) {append after current prefix};

      \draw[ae/mainarrow] (7.7,-1.3) -- (7.7,-2.2);

      % =========================================================
      % Stage 2: manual future candidates
      % =========================================================
      \node[ae/note] at (7.7,-2.7) {manual future candidates};

      % candidate 1
      \node[ae/token] (f11) at (5.3,-3.45) {$z^{(1)}_1$}; \node[ae/token] (f12)
      at (6.5,-3.45) {$z^{(1)}_2$}; \node[ae/token] (f13) at (7.7,-3.45)
      {$z^{(1)}_3$}; \node[ae/token] (f14) at (8.9,-3.45) {$z^{(1)}_4$};
      \node[ae/tinylabel, anchor=west] at (9.9,-3.45) {score $s_1$};

      % candidate 2
      \node[ae/token] (f21) at (5.3,-4.35) {$z^{(2)}_1$}; \node[ae/token] (f22)
      at (6.5,-4.35) {$z^{(2)}_2$}; \node[ae/tinylabel, anchor=west] at
      (9.9,-4.35) {score $s_2$};

      % best candidate
      \node[ae/chosen] (f31) at (5.3,-5.25) {$z^{(*)}_1$}; \node[ae/chosen]
      (f32) at (6.5,-5.25) {$z^{(*)}_2$}; \node[ae/chosen] (f33) at (7.7,-5.25)
      {$z^{(*)}_3$}; \node[ae/tinylabel, anchor=west] at (9.9,-5.25) {highest
      valid score};

      \node[ae/note] at (13.2,-4.35) {candidates scored\\by model likelihood};

      \draw[ae/mainarrow] (7.7,-5.8) -- (7.7,-6.7);

      % =========================================================
      % Stage 3: enforced future tokens
      % =========================================================
      \node[ae/note] at (7.7,-7.15) {best future candidate appended};

      \node[ae/token]   (y1) at (0.0,-7.95) {$x_1$}; \node[ae/token]   (y2) at
      (1.1,-7.95) {$x_2$}; \node[ae/trigger] (y3) at (2.2,-7.95) {$x_3$};
      \node[ae/token]   (y4) at (3.3,-7.95) {$x_4$}; \node[ae/token]   (y5) at
      (4.4,-7.95) {$x_5$}; \node[ae/token]   (y6) at (5.5,-7.95) {$x_6$};
      \node[ae/trigger] (y7) at (6.6,-7.95) {$x_7$};

      \node[ae/chosen]  (y81) at (7.7,-7.95) {$z^{(*)}_1$}; \node[ae/chosen]
      (y82) at (8.9,-7.95) {$z^{(*)}_2$}; \node[ae/chosen]  (y83) at
      (10.1,-7.95) {$z^{(*)}_3$};

      \node[ae/token]   (dots2) at (11.2,-7.95) {$\cdots$};

      \draw[ae/mainarrow] (7.7,-8.4) -- (7.7,-9.3); \node[ae/note] at
      (7.7,-9.85) {advance KV cache\\through forced tokens};

      \draw[ae/mainarrow] (11.75,-7.95) -- (13.0,-7.95); \node[ae/note,
      anchor=west] at (13.1,-7.95) {resume decoding};

    \end{tikzpicture}%
  } \caption{\textbf{Forced future insertion.} When a rule requires a mandatory
    future statement, the controller appends the highest-scoring valid candidate
    and continues decoding from the enforced prefix.}\label{fig:ae_force_future}
\end{figure*}
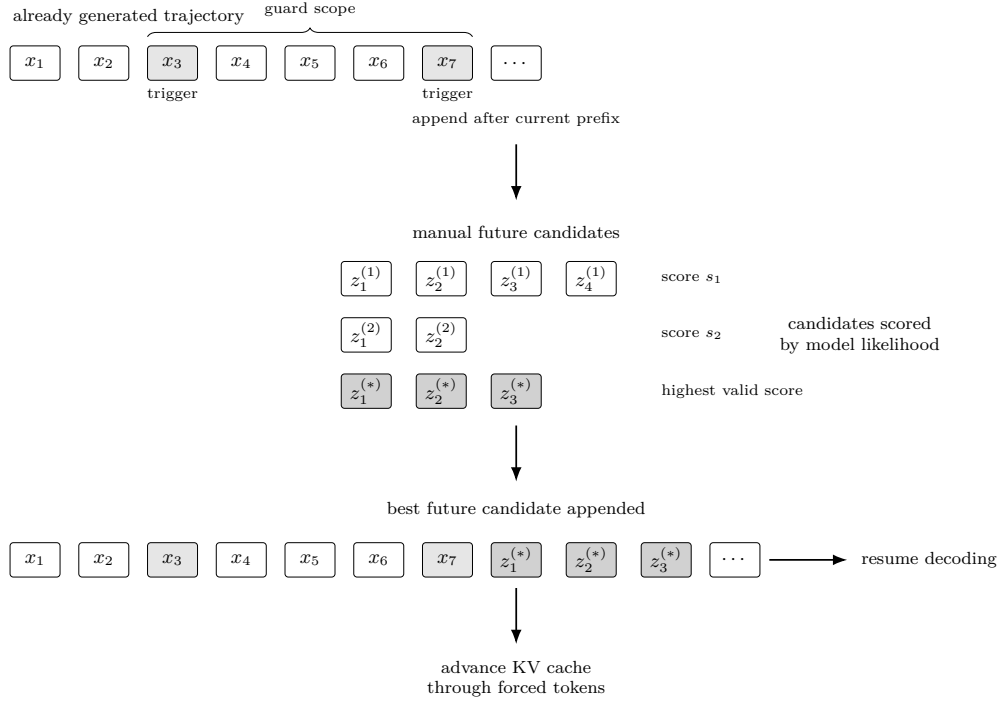

Conceptually, trajectory control allows three classes of intervention,
corresponding to the mechanisms illustrated in
Figures~\ref{fig:ae_retroactive_edit}--\ref{fig:ae_force_future}.

\paragraph{Editing the past (Figure~\ref{fig:ae_retroactive_edit}).} Previously
generated tokens can be replaced when a trigger indicates that a span violates a
protocol constraint. The controller identifies a local edit span within the
guard scope and substitutes it with a manual replacement candidate chosen
according to model likelihood. Because the replacement candidates are defined
by the protocol itself, they are assumed to be valid; the controller therefore
only selects the highest-likelihood candidate before rebuilding the KV cache and
continuing generation.

\paragraph{Selecting among the recently generated continuations
(Figure~\ref{fig:ae_local_rollback}).} When a trigger appears within the
recently generated trajectory, the controller rolls back to the beginning of the
edit scope and explores several alternative continuations generated by the
model. These continuations are ranked by probability, and invalid ones are
rejected according to the protocol rules. The highest-scoring valid continuation
is then accepted. If none of the generated continuations satisfy the constraint,
the controller falls back to predefined manual candidates.

\paragraph{Forcing the future (Figure~\ref{fig:ae_force_future}).} Some protocol
rules require explicit statements or actions to appear in the reasoning
trajectory. Once a trigger condition is detected, the controller selects a
predefined future continuation from several manual candidates and appends it
after the current prefix. The model is then advanced through the forced tokens,
after which normal decoding resumes.

Together, these mechanisms provide a decoding-time control surface for reasoning
trajectories.

\subsection{Local Intervention versus Global Validation}
\textbf{Addresses: F2, F3, F4, F5}

Many reliability approaches rely on \emph{global validation} pipelines, in which
a complete answer is first generated and then verified or refined
\parencite{madaan2023selfrefine,shinn2023reflexion,
gou2024critic,miao2024selfcheck}. Such approaches can substantially improve
overall answer quality, but they often require multiple inference passes and may
discard largely correct trajectories because of a single local mistake.

Trajectory editing instead follows a \emph{local intervention} strategy:
intervene at the violating span while retaining unaffected visible text where
applicable. In this paper, trajectory editing refers to runtime modification of
the generated token sequence to enforce protocol constraints before generation
continues. The unit of control is a \emph{micro-constraint} on admissible local
continuations (representative examples in Appendix~\ref{app:rule_examples}),
which is well aligned with failure modes F4--F5.

\subsection{Micro-Constraints Instead of Expert Decision Trees}
\textbf{Addresses: F4, F5 (and partially F1)}

Traditional rule-based expert systems encode protocols as explicit symbolic rule
structures that can specify large portions of the decision process
\parencite{reddy2022rbes}. While effective in narrow domains, such systems can
become difficult to maintain as rule bases grow in size and complexity
\parencite{chen1994complexity,higa1996maintainability}.

Trajectory editing takes a narrower approach than a full expert system. Instead
of encoding complete reasoning paths, it represents protocols as local
micro-constraints on admissible continuations: forbidding known incorrect
inferences, normalizing protocol-critical terminology, or inserting mandatory
statements once the relevant concept appears. The system therefore constrains
locally dangerous transitions without attempting to encode the entire diagnostic
process.

In certain cases, enforcing such constraints can also improve the consistency
between intermediate reasoning and final conclusions, partially mitigating
unfaithful visible reasoning (F1). For example, if a reasoning trajectory
identifies sudden sensorineural hearing loss (SSNHL), a protocol rule may
enforce urgency language or other protocol-compliance statements associated with
the already generated diagnostic trajectory \textcite{chandrasekhar2019ssnhl}.
Importantly, the rules used in this study do not inject the evaluated management
answer itself (for example, corticosteroid treatment). Instead, diagnostic
concepts act as local trajectory triggers used to prevent unsafe continuations
or protocol omissions. While this does not guarantee full faithfulness of
chain-of-thought reasoning, it helps maintain consistency between visible
reasoning and protocol-critical requirements.

\section{Runtime Framework for Answer Engineering}

Answer Engineering is implemented as a decoding-time runtime controller that
operates alongside standard autoregressive language model inference. The method
does not modify model parameters and does not retrain the underlying
transformer. Instead, it monitors the evolving generation trajectory, detects
rule triggers, proposes candidate trajectory edits, evaluates alternatives under
the model likelihood, and applies the selected edit before generation continues.

Let the language model be parameterized by $\theta$ with next-token distribution
$P_\theta(x_{t+1}\mid x_{1:t})$. Generation therefore proceeds as standard
autoregressive decoding, while the Answer Engineering runtime intervenes only
when protocol constraints are triggered.

The intervention mechanisms themselves were introduced conceptually in
Section~\ref{sec:trajectory_control_principle} and illustrated in
Figures~\ref{fig:ae_retroactive_edit}--\ref{fig:ae_force_future}. The present
section describes how these interventions are executed during decoding.
Implementation details and reproducibility notes are summarized in
Appendix~\ref{app:implementation}. The rule language used to express protocol
constraints is summarized in Appendix~\ref{app:rule_language}, with
representative examples in Appendix~\ref{app:rule_examples}.

\subsection{Rule representation}

Protocol constraints are expressed as declarative rules. Each rule specifies

\begin{itemize}
  \item a \textbf{trigger pattern} that detects a trajectory condition,
  \item a \textbf{scope} identifying the region of the trajectory affected,
  \item a set of \textbf{candidate edits}.
\end{itemize}

Candidate edits represent alternative modifications of the trajectory once the
trigger fires. These edits may include

\begin{itemize}
  \item manually authored protocol phrases (e.g.\ required management
    statements),
  \item locally generated alternative continuations proposed by the model, or
  \item structured edits such as span replacement, rollback continuation, or
    token insertion.
\end{itemize}

Rules therefore encode \emph{local trajectory constraints} rather than complete
reasoning programs. The model remains responsible for generating the reasoning
trajectory, while rules modify only those local segments of the trajectory that
violate protocol constraints.

Rules are authored in a compact Markdown-based domain-specific language that
domain experts can inspect and modify without modifying model code. During
initialization, rules are parsed and compiled into executable runtime plans used
during decoding.

\paragraph{Trajectory edit operator.}

Let the current trajectory be $T = x_{1:t}$. A candidate intervention produces
an edited trajectory

\[
  T' = \mathcal{E}(T;i,j,c) =
  x_{1:i-1} \,\Vert\, c \,\Vert\, x_{j+1:t},
\]

where $[i,j]$ denotes the affected span and $c$ is the candidate replacement
sequence. The edit may correspond to

\begin{itemize}
  \item retroactive span correction,
  \item rollback continuation replacement, or
  \item insertion of a required future statement.
\end{itemize}

Subsequent decoding then proceeds from the edited prefix $T'$ under the same
autoregressive model $P_\theta$.

\subsection{Runtime decoding procedure}

Generation proceeds autoregressively while the runtime monitors the trajectory
for rule triggers. Whenever a trigger fires, the controller evaluates candidate
trajectory edits and applies the highest-scoring compatible intervention before
decoding continues.

All reported experimental configurations use deterministic greedy decoding.
Thus, repeated runs with the same model, dataset, rule set, and environment
produce identical generated trajectories and evaluation outcomes. As a result,
the present evaluation does not involve run-to-run decoding variance, and
interpretation primarily depends on effect magnitude and paired behavioral
changes rather than stochastic generation variability. Formal confidence-
interval estimation was considered but is not emphasized in the main
presentation because the observed effects are substantially larger than
plausible benchmark-level uncertainty and do not depend on fine-grained
discrimination between near-overlapping results.

Runtime is reported as the mean per-case wall-clock time measured over
the full evaluation run, measured from the start of model
decoding after prompt construction. This measurement includes model generation,
rule evaluation, intervention handling, KV-cache reconstruction, and telemetry
construction.

Slowdown is defined as average seconds per case relative to the corresponding
reasoning-generation run within the same diagnostic scope. Values less than
1 indicate faster execution, which corresponds to acceleration rather than
degradation.

Algorithm~\ref{alg:ae_runtime} formalizes the runtime step procedure.

\begin{algorithm}[!htbp]
  \caption{Answer Engineering Runtime Decoding}\label{alg:ae_runtime}

  \small

  \begin{algorithmic}[1]
    \Require model $P_\theta$, prompt $x_{1:n}$, rule plans $\Pi$ \State $T
    \gets x_{1:n}$

    \While{generation not finished} \State sample $x_{t+1}\sim
    P_\theta(\cdot\mid x_{1:t})$ \State append $x_{t+1}$ to $T$ \State
    $\mathcal{G}\gets\mathrm{detect\_triggers}(T,\Pi)$

    \If{$\mathcal{G}\neq\emptyset$} \State bucket $\mathcal{G}$ by earliest edit
    position

    \For{buckets $b$ in ascending edit position} \State
    $\mathcal{C}\gets\mathrm{propose\_candidates}(b,T)$ \State
    $\mathcal{A}\gets\mathrm{filter\_admissible}(\mathcal{C},\Pi)$

    \If{$\mathcal{A}\neq\emptyset$} \State let $i(c)$ be the first affected token position of $c$
    \State compute $s(c;T)=\log P_\theta(c\mid x_{1:i(c)-1})$ \State $\hat{\mathcal{E}}\gets
    \mathrm{select\_compatible}(\mathcal{A},s)$ \State resolve overlaps in
    bucket \State apply $\hat{\mathcal{E}}$ to $T$ deterministically \State $p
    \gets$ earliest modified token \State rebuild KV cache from $p$ \State
    \textbf{break} \EndIf \EndFor \EndIf \EndWhile

    \State \Return $T$
  \end{algorithmic}
\end{algorithm}

The runtime therefore acts as a lightweight trajectory controller embedded
inside standard transformer decoding.

\paragraph{Earliest-bucket execution.}

Triggered rules may affect different portions of the trajectory. To ensure
consistent edits, rules are grouped into \emph{edit buckets} according to the
earliest token position they modify. Buckets are processed in ascending order of
edit position.

Only the earliest bucket that yields admissible edits is executed during a given
decoding step. Later buckets are deferred until generation resumes from the
updated trajectory. This ordering prevents edits from being evaluated on a
trajectory prefix that may subsequently change.

\paragraph{Candidate generation and rollback probing.}

When a bucket is selected, the runtime constructs candidate edits consistent
with the triggered rule semantics. Candidate edits may be

\begin{itemize}
  \item explicit protocol statements provided by the rule definition,
  \item model-generated rollback continuations,
  \item structured edits such as span replacement or insertion.
\end{itemize}

Rollback candidates are generated using a deterministic beam-style probe
procedure applied to the rollback prefix. In the current implementation,
candidate generation uses a shared beam-search routine with optional
grouped-beam diversification and diversity penalties to encourage distinct
continuations before protocol validity filtering.

\paragraph{Candidate scoring.}

Each admissible candidate edit $c$ is scored using the model likelihood under the
prefix preceding its first affected token position $i(c)$:

\[
  s(c;T)=\log P_\theta(c\mid x_{1:i(c)-1}).
\]

For insertions after the current prefix, $i(c)=t+1$, so this reduces to
$\log P_\theta(c\mid T)$. This convention makes explicit that replacement
candidates are scored against the prefix before the affected span, not against
the span being replaced. Using
the model's own scoring function ensures that injected protocol phrases remain
linguistically coherent with the surrounding reasoning.

\paragraph{Conflict resolution and deterministic application.}

Multiple candidate edits may target overlapping spans. The runtime therefore
performs conflict resolution before applying edits. Compatible edits within the
active bucket are selected according to their model scores and applied in
deterministic order.

\paragraph{KV-cache rebuild after edits.}
After a trajectory edit, decoding resumes from the edited prefix by rebuilding
the KV cache from the earliest modified token. Since the cache is an efficiency
device rather than an independent state, this yields the same continuation
distribution as standard decoding from the edited prefix, up to normal numerical
and sampling variance.

\paragraph{External protocol memory.}

Protocol knowledge is maintained externally rather than embedded inside model
parameters. When a rule fires, protocol statements or structured knowledge
fragments can be injected directly into the trajectory as candidate edits.
Because transformer decoding conditions only on the visible prefix, these
inserted tokens immediately become part of the model's effective context.

\subsection{Intervention mechanisms}

During decoding the runtime continuously monitors the evolving trajectory for
rule triggers. When a trigger fires, candidate edits are proposed, evaluated
under the model likelihood, and the selected intervention is applied before
generation resumes.

These runtime interventions implement the three trajectory-control mechanisms
introduced in Section~\ref{sec:trajectory_control_principle}:

\begin{itemize}
  \item \textbf{editing the past} (retroactive span replacement),
  \item \textbf{selecting among recent continuations} (local rollback with
    beam-style probing),
  \item \textbf{forcing the future} (insertion of required protocol statements).
\end{itemize}

\begin{figure}[!htbp]
  \centering
  \begin{tikzpicture}[ font=\small, >=Latex, process/.style={ draw, rounded
        corners=2pt, minimum width=34mm, minimum height=9mm, align=center, inner
        sep=3pt }, decision/.style={ draw, diamond, aspect=2.0, minimum
        width=22mm, minimum height=10mm, align=center, inner sep=1pt },
        arrow/.style={->, thick} ]

    \node[process]  (decode) at (0,0) {Autoregressive decoding}; \node[process]
    (monitor) at (0,-1.6) {Monitor trajectory}; \node[decision] (trigger) at
    (0,-3.4) {Rule triggered?}; \node[process]  (propose) at (0,-5.4) {Construct
    candidate edits}; \node[process]  (score) at (0,-7.0) {Score under model};
    \node[process]  (apply) at (0,-8.6) {Apply selected intervention};
    \node[process]  (cache) at (0,-10.2) {Update prefix / KV cache};

    \draw[arrow] (decode) -- (monitor); \draw[arrow] (monitor) -- (trigger);
    \draw[arrow] (trigger) -- node[right,font=\footnotesize]{yes} (propose);
    \draw[arrow] (propose) -- (score); \draw[arrow] (score) -- (apply);
    \draw[arrow] (apply) -- (cache);

    \draw[arrow] (trigger.east) -- ++(2.0,0)
    node[midway,above,font=\footnotesize]{no} |- (decode.east);

    \draw[arrow] (cache.west) -- ++(-2.0,0) |- (decode.west);

  \end{tikzpicture}
  \caption{
    \textbf{Conceptual runtime control loop.}
    Autoregressive decoding produces tokens while the runtime monitors the
    trajectory for rule triggers. When a rule fires, candidate trajectory edits
    are generated (possibly via beam-style probing), evaluated under the model
    likelihood, and the selected intervention is applied before decoding
    continues from the modified prefix.}\label{fig:ae_pipeline}
\end{figure}
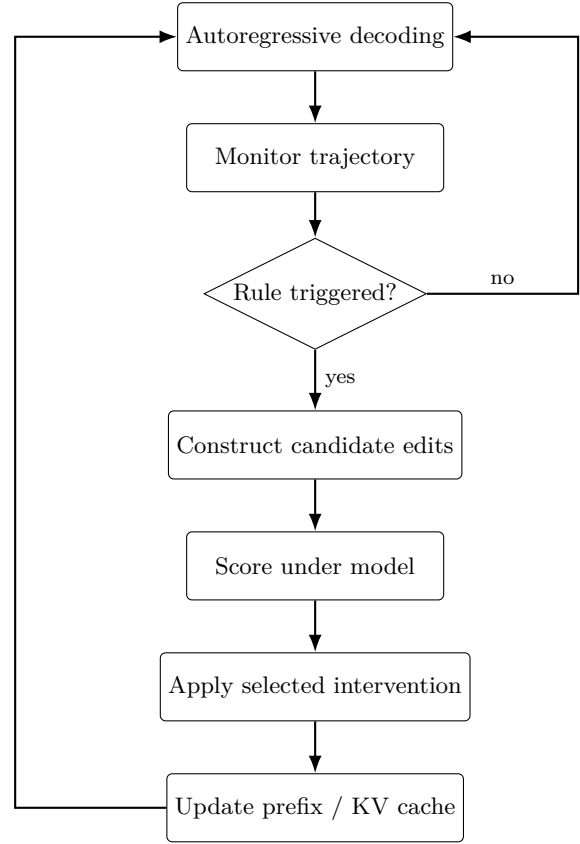

Algorithm~\ref{alg:ae_runtime} specifies the runtime execution semantics, while
Figure~\ref{fig:ae_pipeline} illustrates the same process as a conceptual
control loop.

\section{Experimental Demonstration: Sudden Sensorineural Hearing Loss (SSNHL)}

We evaluate trajectory editing on a diagnostic reasoning task involving sudden
sensorineural hearing loss (SSNHL). The task requires combining multiple
clinical cues and exposes common language-model failure modes such as
inconsistent interpretation of diagnostic tests, premature conclusions, and
omission of mandatory protocol steps.

All experiments used the publicly released medical language model
\hfmodel{OpenMeditron/Meditron3-8B} % chktex 8
at revision
\href{https://huggingface.co/OpenMeditron/Meditron3-8B/tree/15914bcb040cd1a4f263afcd85b84f09ad2efd95}
{\texttt{15914bcb040c}}~\parencite{openmeditron2025meditron3_8b}. No local
model modifications were applied. All evaluations were executed with
deterministic decoding settings, ensuring that differences between
configurations reflect intervention behavior rather than sampling variability.
Trajectory editing was applied during decoding using the runtime control loop in
Figure~\ref{fig:ae_pipeline}.

The goal of this experiment is to test feasibility of decoding-time trajectory
control on a narrow, protocol-sensitive benchmark rather than to model the full
complexity of clinical otologic decision-making.

\subsection{Clinical background}

Sudden sensorineural hearing loss (SSNHL) is a time-sensitive clinical condition
requiring prompt evaluation.\footnote{Clinically, SSNHL is typically defined as
a sensorineural hearing loss of at least 30 dB affecting three consecutive
frequencies and developing within 72 hours \parencite{chandrasekhar2019ssnhl}.}
In this benchmark, diagnosis depends on combining symptom timing with bedside
Weber and Rinne tuning-fork findings under standard AAO-HNS interpretation
rules.\footnote{Appendix~\ref{app:clinical_endpoint} reproduces the
benchmark-facing interpretation summary used for Weber and Rinne testing.}

\subsection{Dataset}
We evaluate on a public deterministic dataset of synthetic, parameterized
clinical vignettes designed to isolate a protocol-sensitive decision boundary
while preserving a clinically plausible presentation structure. The design
intentionally uses symmetric paired templates for SSNHL-consistent and
conductive cases: age, laterality, time since onset, distractor details, and
surface wording vary, while the decisive Weber/Rinne and otoscopic findings are
changed in a class-consistent way. For evaluation, we use two balanced subsets:
$N=\PaperEvalN$ SSNHL-consistent cases and $N=\PaperEvalN$ contrastive
conductive cases. Each case combines symptom timing, Weber/Rinne findings, and
class-aligned otoscopic examination, with secondary distractor details included
but not made decisive. This benchmark is used for controlled relative assessment
of reasoning trajectories rather than as a surrogate for full-spectrum clinical
performance or natural clinical prevalence, and the detailed template design
rationale is provided in Appendix~\ref{app:clinical_endpoint}.

The synthetic generator also functions as a controlled perturbation mechanism.
Each template family is rendered under seeded label-preserving perturbations,
including surface-wording variation, laterality-preserving paired substitutions,
numeric variation in age and time since onset, evidence-order variation in
symptom lists, and inclusion or removal of non-decisive distractor details.
These perturbations preserve the benchmark label because the decisive
Weber/Rinne pattern and class-aligned otoscopic findings remain fixed within
each diagnostic branch. The present evaluation therefore tests intervention
behavior across controlled surface and distractor variation, but does not
separately report perturbation-sensitivity metrics by perturbation type.
Ambiguity-introducing perturbations, where the clinical label is intentionally
weakened or made uncertain, are left for future work.

\subsection{Evaluation criterion}

For benchmarking purposes, protocol adherence is operationalized as a binary
endpoint: whether the answer explicitly recommends steroid therapy. This endpoint
captures the key treatment branch distinguishing suspected SSNHL from conductive
hearing loss.\footnote{Appendix~\ref{app:clinical_endpoint} explains the endpoint
choice and contrasts it with alternative candidate signals.} For SSNHL cases,
answers that explicitly recommend steroid therapy satisfy the benchmark
criterion; for conductive cases, they do not. This endpoint is intentionally
narrow and isolates the key therapeutic branch in the protocol, testing
adherence to a decisive management step rather than full clinical answer
quality.

Each generated case is treated as an independent sample in metric
computation.

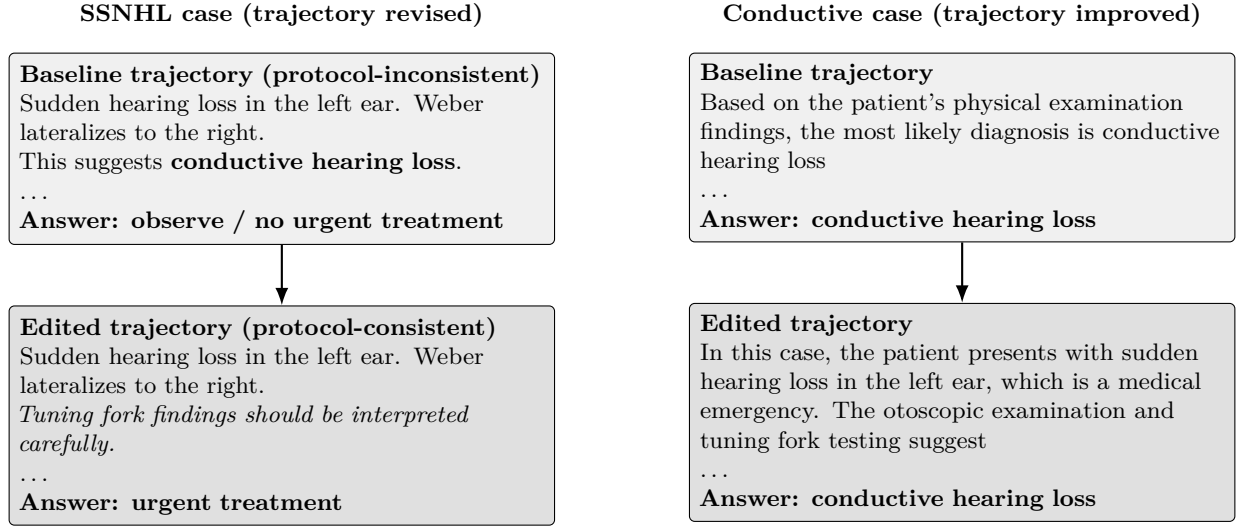
\begin{figure*}[!htbp]
  \centering
  \begin{tikzpicture}[ font=\small, >=Latex, box/.style={ draw, rounded
      corners=2pt, align=left, inner sep=4pt, text width=0.42\linewidth },
      bad/.style={box, fill=black!6}, good/.style={box, fill=black!12},
      arrow/.style={->, thick}, title/.style={font=\small\bfseries} ]

    \node[title] at (-4.5,0.5) {SSNHL case (trajectory revised)}; \node[title]
    at (4.5,0.5) {Conductive case (trajectory improved)};

    \node[bad, anchor=north] (b1) at (-4.5,0) { \textbf{Baseline trajectory
      (protocol-inconsistent)}\\
      Sudden hearing loss in the left ear. Weber lateralizes to the right.\\
      This suggests \textbf{conductive hearing loss}.\\
      \dots\\
      \textbf{Answer: observe / no urgent treatment}
    };

    \node[good, anchor=north] (e1) at ([yshift=-8mm]b1.south) { \textbf{Edited
      trajectory (protocol-consistent)}\\
      Sudden hearing loss in the left ear. Weber lateralizes to the right.\\
      \emph{Tuning fork findings should be interpreted carefully.}\\
      \dots\\
      \textbf{Answer: urgent treatment}
    };

    \draw[arrow] (b1.south) -- (e1.north);

    \node[bad, anchor=north] (b2) at (4.5,0) { \textbf{Baseline trajectory}\\
      Based on the patient's physical examination findings, the most likely
      diagnosis is conductive hearing loss\\
      \dots\\
      \textbf{Answer: conductive hearing loss}
    };

    \node[good, anchor=north] (e2) at ([yshift=-8mm]b2.south) { \textbf{Edited
      trajectory}\\
      In this case, the patient presents with sudden hearing loss in the left
      ear, which is a medical emergency. The otoscopic examination and tuning
      fork testing suggest\\
      \dots\\
      \textbf{Answer: conductive hearing loss}
    };

    \draw[arrow] (b2.south) -- (e2.north);

  \end{tikzpicture}

  \caption{ \textbf{Two representative trajectories.}\\
    \textbf{Left:} trajectory editing revises an SSNHL case by enforcing
    protocol-consistent interpretation of tuning fork findings, leading to
    protocol-consistent management.\\
    \textbf{Right:} in conductive hearing loss cases, trajectory editing can
    also improve reasoning fidelity to the stem, preserving the conductive
    diagnosis while reducing protocol-inconsistent detours.
    }\label{fig:ssnhl_examples}
\end{figure*}

Figure~\ref{fig:ssnhl_examples} illustrates how trajectory revision can improve
protocol-consistent reasoning and management.

\subsection{Overall results}\label{sec:ssnhl_results}

\begin{table*}[!htbp]
  \centering
  \small
  \begin{tabular}{lccccc}
    \toprule
    & \multicolumn{2}{c}{SSNHL} & \multicolumn{2}{c}{Conductive} &
    \multirow{2}{*}{Balanced accuracy} \\
    \cmidrule(lr){2-3}\cmidrule(lr){4-5} Method & Accepted & Slowdown & Accepted
    & Slowdown & \\
    \midrule
    Baseline LLM & \SSNHLBaselineAcceptedPct{} & \SSNHLBaselineSlowdownX{} &
    \ConductiveBaselineAcceptedPct{} & \ConductiveBaselineSlowdownX{} &
    \CombinedBaselineBalancedAccuracyPct{} \\
    Reasoning & \SSNHLReasoningAcceptedPct{} & \SSNHLReasoningSlowdownX{} &
    \ConductiveReasoningAcceptedPct{} & \ConductiveReasoningSlowdownX{} &
    \CombinedReasoningBalancedAccuracyPct{} \\
    Global validation & \SSNHLGlobalValidationAcceptedPct{} &
    \SSNHLGlobalValidationSlowdownX{} & \ConductiveGlobalValidationAcceptedPct{}
    & \ConductiveGlobalValidationSlowdownX{} &
    \CombinedGlobalValidationBalancedAccuracyPct{} \\
    Local validation & \SSNHLLocalValidationAcceptedPct{} &
    \SSNHLLocalValidationSlowdownX{} & \ConductiveLocalValidationAcceptedPct{} &
    \ConductiveLocalValidationSlowdownX{} &
    \CombinedLocalValidationBalancedAccuracyPct{} \\
    Replace + After only & \SSNHLReplaceAfterAcceptedPct{} &
    \SSNHLReplaceAfterSlowdownX{} & \ConductiveReplaceAfterAcceptedPct{} &
    \ConductiveReplaceAfterSlowdownX{} &
    \CombinedReplaceAfterBalancedAccuracyPct{} \\
    Trajectory editing & \SSNHLTrajectoryAcceptedPct{} &
    \SSNHLTrajectorySlowdownX{} & \ConductiveTrajectoryAcceptedPct{} &
    \ConductiveTrajectorySlowdownX{} & \CombinedTrajectoryBalancedAccuracyPct{}
    \\
    \bottomrule
  \end{tabular}
\caption{Intervention-policy ablation matrix for the implemented policy subset
  on SSNHL ($N=\PaperEvalN{}$) and contrastive conductive cases
  ($N=\PaperEvalN{}$). Slowdown is reported relative to the reasoning run within
  the same diagnostic scope; values below $1\times$ indicate acceleration.
  Balanced accuracy averages the SSNHL and conductive acceptance rates when both
  scopes are available. The reported subset includes unguided generation,
  reasoning, global validation, local validation, Replace/After-only editing,
  and full trajectory editing. Trajectory editing combines local validation with
  Replace and After rules into a coordinated intervention
  mechanism.}\label{tab:ssnhl-results}
\end{table*}

Table~\ref{tab:ssnhl-results} summarizes performance across progressively more
localized intervention mechanisms applied to the generation process. The
evaluated configurations range from prompt-level adjustments to trajectory-level
coordination of validation and repair.

Table~\ref{tab:ssnhl-results} should be interpreted as the intervention-policy
ablation matrix for the currently implemented policy set. It reports a
documented subset of possible intervention policies: unguided generation,
reasoning, global validation, local validation, Replace/After-only editing, and
the full trajectory-editing stack. Branch-aware control and adaptive or
confidence-triggered policies are discussed in Future Work
(Sections~\ref{sec:future_branch_aware} and~\ref{sec:future_adaptive_policies})
and are therefore not included in the present ablation matrix.

The ruleset was finalized before test evaluation, and no rule modifications were
made after examining test outputs except for post hoc failure analysis reported
in the paper.

\paragraph{Baseline LLM}

The unguided baseline exhibits a strong diagnostic bias toward SSNHL treatment.
In this configuration, the model frequently recommends steroid therapy
regardless of the clinical presentation, producing moderate acceptance rates on
SSNHL cases but near-zero acceptance on conductive contrast cases. This pattern
indicates a prior-dominated response strategy rather than reliable differential
reasoning, and results in low balanced accuracy.

\paragraph{Reasoning}

Introducing step-by-step reasoning changes the error pattern but does not
resolve the underlying instability. Compared with the unguided baseline, the
reasoning configuration increases acceptance on conductive cases but reduces
acceptance on true SSNHL cases. Balanced accuracy shifts relative to the
baseline, but this change reflects redistribution of errors rather than
consistent protocol compliance. The model becomes less biased toward SSNHL but
more likely to misclassify true SSNHL cases as conductive hearing loss.

\paragraph{Global validation}

Global validation is a widely used constraint-enforcement strategy in sequence
generation systems, in which multiple candidate responses are generated and
filtered against protocol requirements before final selection
\parencite{hokamp2017constrained,zhang2023constrained,koo2024automata}. In the
present evaluation, alternative responses are produced using group beam search
to explore diverse reasoning trajectories during decoding
\parencite{vijayakumar2016diverse}. Candidate outputs that violate critical
constraints are rejected or replaced with compliant alternatives.

This mechanism stabilizes protocol adherence relative to the reasoning
configuration by reducing the number of unacceptable decisions while preserving
the underlying generation process. Because validation operates at the level of
completed candidate outputs, the reasoning trajectory itself is not modified.
Observed adherence gains therefore arise from selecting compliant alternatives
rather than from changing the model's diagnostic bias during generation.

\paragraph{Local validation}

Local validation extends this validation paradigm by applying the same
constraint checks earlier in the generation process, monitoring reasoning steps
as they are produced rather than evaluating only completed responses. Instead of
generating full alternative outputs and selecting among them, local validation
detects protocol violations at intermediate points in the trajectory and can
trigger corrective actions without regenerating the entire response.

In the present evaluation, this earlier detection does not produce a substantial
shift in diagnostic acceptance rates relative to global validation, indicating
that the timing of validation alone does not materially alter the underlying
decision bias of the model. However, local validation provides a clear
operational advantage: violations can be identified and addressed incrementally,
resulting in reduced regeneration overhead and improved runtime efficiency
compared with output-level validation.

\paragraph{Manual Replace + After only}

Replace and After interventions provide a direct-edit alternative to
validation-based selection. Instead of generating multiple candidate
trajectories and filtering them against protocol constraints, these rules
enforce manually authored protocol-compatible trajectory fragments at specific
trigger points. When several candidate fragments are available, they are scored
under the model likelihood, and the highest-likelihood compatible fragment is
inserted so that the edited trajectory remains linguistically coherent with the
surrounding generation.

This mechanism can correct isolated protocol failures without regenerating or
selecting among full candidate responses. However, when applied without local
validation, edits are limited to predefined trigger-repair patterns and do not
provide a general mechanism for detecting and resolving arbitrary local
violations during generation.

\paragraph{Trajectory editing (Local validation + Replace + After)}

Trajectory editing combines local validation with coordinated Replace and After
rules to detect, repair, and propagate corrections during generation. In this
configuration, acceptance rates increase simultaneously in both diagnostic
scopes rather than shifting between them. Balanced accuracy rises substantially
relative to all baseline and intermediate configurations, indicating that
coordinated local correction of reasoning steps can improve decision quality
without introducing compensatory degradation in other task variants.

Importantly, the applied rules were designed to enforce faithful
protocol-compliant reasoning rather than to promote a specific diagnosis. For
example, early diagnostic conclusions were prohibited uniformly, including
premature inference of both conductive hearing loss and SSNHL.\@ All
trajectories were required to follow the same reasoning sequence, beginning with
structured test interpretation and proceeding to diagnosis only after required
evidence had been evaluated. This symmetric enforcement ensures that observed
adherence gains do not arise from suppressing alternative diagnostic pathways or
artificially increasing the relative probability of one class.

Under these constraints, simultaneous adherence gains observed in both SSNHL and
contrastive conductive cases are consistent with stabilization of the reasoning
process rather than simple redistribution of diagnostic bias. Enforcing
consistent reasoning order is consistent with reduced premature conclusions and
more stable alignment between intermediate reasoning steps and final decisions
across task variants. This contrastive adherence effect is particularly
informative because a rule set that merely favored or injected the SSNHL
endpoint would be expected to degrade conductive cases rather than improve them.

Across configurations, protocol-constrained behavior becomes progressively more
stable as intervention mechanisms become more localized and coordinated. This
pattern is consistent with many observed incorrect decisions being compatible
with premature or inconsistent reasoning trajectories rather than missing
medical knowledge alone. In this controlled benchmark, stabilizing these
trajectories during decoding restores protocol-consistent outcomes in a large
fraction of cases.

\subsection{Runtime telemetry and intervention behavior}

Tables~\ref{tab:runtime-interventions} and~\ref{tab:runtime-alternatives}
summarize trajectory-editing behavior on the SSNHL subset only
($N=\PaperEvalN$).

Avoid-rule resolution accounts for the majority of accepted edits in this run.
Alternative trajectory generation introduces additional decoding work but is
typically resolved after evaluating only a small number of candidate
continuations before generation resumes.

Table~\ref{tab:runtime-alternatives} characterizes the search depth required to
resolve avoid-rule episodes. Percentile values indicate the minimum number of
alternative trajectories required to resolve the specified fraction of episodes.
The ``episodes exhausting alternative budget'' metric identifies episodes in
which the predefined alternative budget was fully consumed%
\footnote{The ``episodes exhausting alternative budget'' metric identifies
episodes in which the predefined alternative budget was fully consumed. Budget
exhaustion does not indicate failure to produce a final answer; in such cases,
generation continues using the highest-scoring available trajectory after the
alternative search budget is reached.}

All telemetry metrics are computed directly from structured runtime logs using
the same instrumentation used to generate the reported evaluation tables,
ensuring consistency between behavioral telemetry and outcome metrics.

\begin{table}[!htbp]
  \centering
  \small
  \setlength{\tabcolsep}{4pt}

  \begin{tabular}{lc}
    \toprule
    Metric & Value \\
    \midrule

    ``Avoid'' interventions per case & \RuntimeAvoidInterventionsPerCase{} \\

    All interventions per case & \RuntimeAvgInterventionsPerCase{} \\

    \bottomrule
  \end{tabular}

  \caption{ Intervention frequency statistics computed over all evaluated cases
    on the SSNHL subset ($N=\PaperEvalN$). Values are averaged across all cases,
    including cases with zero interventions. }\label{tab:runtime-interventions}

\end{table}

\begin{table}[!htbp]
  \centering
  \small
  \setlength{\tabcolsep}{4pt}

  \begin{tabular}{lc}
    \toprule
    Metric & Value \\
    \midrule

    Average alternatives tried & \RuntimeAvgAlternativesTried{} \\

    Alternatives for 50\% resolution &
    \RuntimeAlternativesForFiftyPctResolution{} \\

    Alternatives for 80\% resolution &
    \RuntimeAlternativesForEightyPctResolution{} \\

    Episodes exhausting alternative budget & \RuntimeRanOutOfAlternatives{} \\

    \bottomrule
  \end{tabular}

  \caption{ Alternative-search behavior during avoid-rule resolution on the
    SSNHL subset ($N=\PaperEvalN$). Percentile rows indicate the minimum number
    of alternative trajectories required to resolve the specified fraction of
    avoid-rule episodes. }\label{tab:runtime-alternatives}

\end{table}

\subsection{Edit trace examples}

Table~\ref{tab:edit-traces} provides representative edit traces from runtime
logs, including protocol-relevant interpretations of Weber and Rinne findings
and the corresponding management statement.

\begin{table*}[!htbp]
  \centering
  \small
  \begin{tabular}{p{5cm} >{\raggedright\arraybackslash}p{4cm} p{5cm}}
    \toprule
    Baseline fragment & Rule trigger & Edited fragment \\
    \midrule

    ``The patient presents with sudden hearing loss in the \textbf{left} ear,
    which is a concerning symptom. The \textbf{Weber} test lateralizing to the
    \textbf{right} ear suggests that the left ear has a
    \textbf{conductive~\failmark}'' & Contralateral Weber lateralization
    invalidates conductive inference & ``The patient presents with sudden
    hearing loss in the left ear, which is a concerning symptom. The Weber test
    lateralizes to the right ear\emph{, indicating that the left ear is worse,
    and the Rinne test is positive on the left ear\ldots}'' \\

    \addlinespace

    ``\ldots{} and the \textbf{Rinne} test is \textbf{positive} on the left ear,
    suggesting that the hearing loss is \textbf{conductive~\failmark}'' & Positive
    Rinne invalidates conductive inference & ``\ldots{}, and the Rinne test is
    positive on the left ear\emph{, indicating that air conduction is better
    than bone conduction\ldots}'' \\

    \addlinespace

    ``The patient presents with \textbf{sudden} hearing loss \ldots These
    findings are consistent with \textbf{sensorineural hearing loss}'' & Enforce
    ``sudden'' qualifier for sensorineural hearing loss & ``The patient presents
    with sudden hearing loss \ldots These findings are consistent with
    \emph{\textbf{sudden}} sensorineural hearing loss\ldots{}'' \\

    \addlinespace

    ``\ldots{}These findings are consistent with \textbf{sudden sensorineural
    hearing loss}'' & Add urgent treatment statement for SSNHL & ``\ldots{}These
    findings are consistent with sudden sensorineural hearing loss.\ \emph{This
    condition requires urgent treatment\ldots{}}'' \\

    \bottomrule
  \end{tabular}

  \caption{ Representative edit traces extracted from runtime logs.
  }\label{tab:edit-traces}
\end{table*}

\subsection{Paired trajectory changes across target and contrastive tasks}

The conductive task serves as a contrastive control condition for evaluating
unintended side effects of protocol enforcement while also providing an
independent signal about trajectory behavior outside the target protocol.

\begin{table*}[!htbp]
  \centering
  \small
  \begin{tabular}{llccc}
    \toprule
    Method & Scope & Improved & Degraded & $\Delta$ acc. \\
    \midrule
    Global validation & SSNHL & \SSNHLGlobalValidationImprovedCount{} &
    \SSNHLGlobalValidationDegradedCount{} & \SSNHLGlobalValidationDeltaPP{} \\
    Local validation & SSNHL & \SSNHLLocalValidationImprovedCount{} &
    \SSNHLLocalValidationDegradedCount{} & \SSNHLLocalValidationDeltaPP{} \\
    Replace + After only & SSNHL & \SSNHLReplaceAfterImprovedCount{} &
    \SSNHLReplaceAfterDegradedCount{} & \SSNHLReplaceAfterDeltaPP{} \\
    Trajectory editing & SSNHL & \SSNHLTrajectoryImprovedCount{} &
    \SSNHLTrajectoryDegradedCount{} & \SSNHLTrajectoryDeltaPP{} \\
    Trajectory editing & Conductive & \ConductiveTrajectoryImprovedCount{} &
    \ConductiveTrajectoryDegradedCount{} & \ConductiveTrajectoryDeltaPP{} \\
    \bottomrule
  \end{tabular}
  \caption{Paired change summary across SSNHL and conductive runs, reported in
raw counts. Improvements substantially exceed degradations across reported
settings; the observed directional asymmetry would correspond to highly
significant paired comparisons ($p < 0.001$), although formal paired
significance testing is not emphasized in the present
evaluation.}\label{tab:paired-trajectory-changes}
\end{table*}

Table~\ref{tab:paired-trajectory-changes} summarizes the paired changes induced
by the different intervention bundles across both tasks. The SSNHL rows show
that the full trajectory-editing stack delivers the largest net gain on the
target protocol. In the conductive setting, trajectory editing also produces a
substantial positive shift (\ConductiveTrajectoryDeltaPP{}), with improvements
(\ConductiveTrajectoryImprovedPP{}) exceeding degradations
(\ConductiveTrajectoryDegradedPP{}).

Because Table~\ref{tab:paired-trajectory-changes} reports paired improved-
versus-degraded outcomes directly, interpretation primarily depends on the
direction and magnitude of paired changes rather than on marginal acceptance
rates alone. Across the reported comparisons, improvements substantially
exceed degradations, indicating a consistent directional effect of
intervention. Formal paired significance testing and exact p-value reporting
are therefore not emphasized here, since the observed asymmetry is already
large and directionally unambiguous.

We treat the degraded rows in
Table~\ref{tab:paired-trajectory-changes} as collateral intervention cases:
outputs that were correct under the comparison run but became incorrect after
intervention. Manual review of the
\SSNHLTrajectoryDegradedCount{} degraded SSNHL trajectory-editing cases found
that these failures were not cleanly attributable to a single rule. In the
reviewed cases, the reasoning baseline often reached or strongly implied a
diagnosis before explicitly analyzing Weber and Rinne findings, and several
responses mentioned these findings primarily as post-diagnosis rationalization
rather than as premises from which the diagnosis was derived.

This review suggests that even under reasoning-oriented prompting, the
evaluated model did not always maintain strong step-wise reasoning fidelity. In
some reviewed cases, reasoning-form outputs remained compressed or
answer-first despite explicit step-by-step instructions. Trajectory editing
therefore often operated by repairing already premature reasoning rather than
refining fully step-wise analysis.

The reviewed collateral cases fell into three recurring categories:

\begin{enumerate}[leftmargin=*,nosep]
\item \textbf{Management-endpoint failures.}
Trajectory editing sometimes repaired reasoning order while the final
management remained referral-first rather than explicitly recommending steroid
treatment.

\item \textbf{Diagnostic reinterpretation instability.}
Repair sometimes destabilized interpretation of tuning-fork findings,
producing conductive, otitis-media, or vascular explanations despite
SSNHL-consistent findings.

\item \textbf{Intermediate-reasoning instability.}
Repeated local correction sometimes produced valid-looking intermediate
reasoning without preserving the benchmark-critical management endpoint.
\end{enumerate}

These observations help contextualize the degraded cases reported in
Table~\ref{tab:paired-trajectory-changes}. The presence of collateral failures
does not imply that intervention merely forces target labels or that failures
are cleanly reducible to individual rules. Instead, the intervention mechanism
appears to perform genuine trajectory modification, sometimes stabilizing
reasoning and sometimes interacting with already unstable or answer-first
trajectories. Consistent with this interpretation, local trajectory
interventions improve not only protocol compliance on the SSNHL task but also
clinical alignment in cases governed by a different diagnostic pathway.

At the same time, the result highlights an important interpretive limitation:
the current evaluation does not fully disentangle gains arising from
task-specific protocol repair versus broader trajectory stabilization.

\subsection{When trajectory repair fails}\label{sec:repair_failures}

The collateral cases discussed above suggest a broader failure pattern extending
beyond isolated degradation events. We therefore manually reviewed both
\emph{repair failures} (where the reasoning baseline already failed and
trajectory editing did not recover correctness) and \emph{collateral
degradations} (where the reasoning baseline was correct but intervention
introduced failure).

The reviewed failures were drawn directly from repository evaluation artifacts
and runtime logs generated by the same deterministic reproduction pipeline used
for
Tables~\ref{tab:runtime-interventions},~\ref{tab:runtime-alternatives},~\ref{tab:edit-traces},
and~\ref{tab:paired-trajectory-changes}. The repair traces shown below therefore
represent reproducible runtime behavior rather than manually constructed
examples.

Across these cases, a recurring mechanism emerged: the model frequently
committed prematurely to a diagnosis before completing the intended step-wise
analysis. Trajectory editing therefore often operated by repairing already
premature reasoning rather than refining naturally step-wise trajectories.

In successful cases, intervention redirected generation toward
protocol-compliant reasoning. However, difficult cases frequently exhibited
\emph{persistent diagnostic commitment}: the model repeatedly returned to early
diagnostic closure despite local repair. AE therefore did not simply miss
violations. Instead, the runtime often detected and repaired them repeatedly,
while the underlying generation policy continued attempting diagnosis-first
reasoning.

\begin{table}[t]
\centering
\small
\begin{tabularx}{\columnwidth}{>{\raggedright\arraybackslash}p{0.20\columnwidth}X>{\raggedright\arraybackslash}p{0.18\columnwidth}}
\toprule
Attempt & Visible trajectory & Runtime response \\
\midrule
Initial &
``Based on the patient's presentation, \textbf{the most likely diagnosis is sudden sensorineural} \ldots'' &
rewritten \\
Attempt 2 &
``Based on the patient's history and examination findings, \textbf{the most likely diagnosis is sudden sensorineural} \ldots'' &
rewritten \\
Attempt 3 &
``Based on the patient's history and physical examination findings, \textbf{the most likely diagnosis is sudden sensorineural} \ldots'' &
rewritten \\
Attempt 4 &
``In this case, the patient presents with \ldots'' &
non-diagnostic rewrite \\
\bottomrule
\end{tabularx}
\caption{Representative repeated repair cycle. The model repeatedly returns to
premature SSNHL commitment despite successive rewrites delaying
diagnosis.}\label{tab:repair-cycle}
\end{table}

Table~\ref{tab:repair-cycle} illustrates a representative repair pattern. The
model repeatedly attempted to commit immediately to SSNHL before completing
step-wise analysis. Each time, AE rewrote the continuation, yet generation
returned to essentially the same diagnostic commitment before producing a
temporarily non-diagnostic opening.

Importantly, this did not mean that the trajectory had become protocol
compliant. Softening the opening could delay early diagnosis without eliminating
the underlying tendency to interpret findings prematurely.

\begin{table}[t]
\centering
\small
\begin{tabularx}{\columnwidth}{>{\raggedright\arraybackslash}p{0.22\columnwidth}X>{\raggedright\arraybackslash}p{0.22\columnwidth}}
\toprule
Stage & Visible trajectory & Runtime response \\
\midrule
Opening softened &
``In this case, the patient presents with sudden hearing loss \ldots'' &
not sufficient \\
Premature interpretation returns &
``The tuning fork testing suggests that the hearing loss is conductive'' &
avoid triggered \\
Repair applied &
``\#\#\# Clinical Assessment The patient is a'' &
trajectory restarted \\
Renewed structured reasoning &
``The key findings include \ldots Tuning Fork Testing \ldots indicating a conductive'' &
another avoid triggered \\
\bottomrule
\end{tabularx}
\caption{Softening the opening did not satisfy the protocol. The model still
returned to diagnostic interpretation before adequate Weber/Rinne analysis, so
AE continued repairing the trajectory.}\label{tab:repair-persistence}
\end{table}

Table~\ref{tab:repair-persistence} illustrates a second pattern. Even after the
opening was rewritten into less diagnosis-first language, the model quickly
returned to premature interpretation of the tuning-fork findings. The repair
therefore delayed diagnostic closure but did not by itself induce stable
step-wise reasoning.

These observations suggest that many AE failures arise not primarily from missed
violations or isolated rule errors but from persistent diagnosis-first
generation dynamics. The model often stabilizes around an early diagnostic
hypothesis and repeatedly attempts to return to that hypothesis despite
intervention.

This pattern produced several characteristic outcomes:

\begin{enumerate}[leftmargin=*,nosep]
\item \textbf{Repair failure.}
The reasoning baseline already failed and repeated intervention did not recover
the correct endpoint.

\item \textbf{Collateral degradation.}
The reasoning baseline reached a correct endpoint, but repeated repair
destabilized interpretation and displaced the original diagnosis or management
decision.

\item \textbf{Commitment persistence.}
Intervention was repeatedly detected and applied, yet the model continued
attempting premature diagnostic closure and failed to sustain genuinely
step-wise reasoning.
\end{enumerate}

These findings suggest an important boundary condition of visible trajectory
repair. AE can redirect or repair generated reasoning, but when the underlying
generation policy repeatedly violates the intended reasoning order, local repair
alone may be insufficient to induce stable step-wise analysis.

\subsection{Model Sensitivity and Baseline
Pathologies}\label{sec:model_sensitivity}

Although all formal results in this study use
\hfmodel{OpenMeditron/Meditron3-8B}, we conducted exploratory pilot % chktex 8
runs on additional medical-language models to assess how trajectory-level
control behaves under different baseline reasoning characteristics. In
particular, we tested \hfmodel{OpenMeditron/Meditron3-Qwen2.5-7B} and % chktex 8
\hfmodel{HPAI-BSC/Llama3.1-Aloe-Beta-8B}. These pilots were performed on small
samples far smaller than the primary benchmark and were intended for qualitative
inspection rather than statistical comparison. For this reason, we report
behavioral patterns rather than specific percentages.

The Qwen-based model was substantially slower in our setup but exhibited the
same qualitative improvement pattern as the primary model: local interventions
and full trajectory editing improved reasoning outcomes relative to baseline.
This observation suggests that the core mechanism of trajectory-level
intervention transfers across related architectures when the model maintains
interpretable intermediate reasoning steps.

In contrast, the Aloe model exhibited a baseline pathology rather than a
trajectory-level failure. Before any intervention, it frequently defaulted to an
SSNHL diagnosis across diverse presentations, including conductive cases.
Inspection of generated reasoning traces showed that the model often misread
Weber/Rinne findings early in the reasoning process and committed immediately to
the emergency diagnosis regardless of case-specific evidence. As a result, the
baseline appeared highly accurate for SSNHL cases but performed poorly on
contrastive conductive cases.

This behavior indicates that some models may enter evaluation with strongly
skewed diagnostic priors or limited reasoning fidelity, producing superficially
high performance on a subset of tasks while failing to perform meaningful
differential diagnosis.

\subsection{Large Gains and Trajectory Instability in Weak
Models}\label{sec:trajectory_instability}

Given the baseline pathology described in Section~\ref{sec:model_sensitivity},
in which the Aloe model frequently defaulted to an SSNHL diagnosis regardless of
case-specific evidence, we constructed a rule configuration in which conductive
hearing loss served as the primary condition while SSNHL remained the
contrastive case. In this setting, trajectory editing produced substantial
improvements relative to baseline, often increasing accuracy by several-fold.
These gains were qualitatively similar to the main experiments but were more
pronounced because the initial model performance was substantially lower.

However, the intervention dynamics differed from those observed with the primary
model. Because the baseline reasoning trajectory frequently committed early to
an incorrect diagnosis, redirecting the trajectory often required repeated
fallback corrections within the same generation. The frequency of such
corrections was substantially higher than in the primary experiments, where
fallback interventions occurred only intermittently.

When fallback density became high, the resulting responses sometimes exhibited
unstable reasoning structure, including excessive verbosity, repetition, or
internally inconsistent phrasing. These effects did not typically alter the
final protocol-compliant decision but indicated that the model had been driven
into low-probability internal states through repeated trajectory redirection. In
other words, large improvements in final-answer accuracy did not necessarily
correspond to improvements in reasoning quality or stability.

This observation suggests that trajectory-level intervention remains capable of
substantially improving accuracy even for weak or poorly calibrated models, but
that the efficiency and robustness of the resulting reasoning depend on the base
model's ability to sustain coherent trajectories under repeated correction.

More broadly, these findings motivate a separate line of investigation into the
faithfulness and controllability of model reasoning processes. In particular,
future work should examine how readily a model can diverge from an incorrect
early diagnosis, whether alternative reasoning trajectories remain accessible
during generation, and how diverse candidate continuations can be produced at
critical pivot points. Such studies may involve systematic comparison of probe
generation strategies, including approaches that explicitly encourage trajectory
diversity (e.g., self-training or iterative refinement methods that generate
diverse reasoning paths, such as STaR~\parencite{zelikman2022star} or
self-consistency decoding~\parencite{wang2023selfconsistency}), and may provide
practical criteria for selecting models that are most responsive to
trajectory-level intervention.

\section{Limitations}

The present evaluation is conducted within a controlled, single-protocol
diagnostic boundary designed to isolate specific reasoning failures. The
benchmark constrains the decision space to a narrow causal structure in which
protocol violations are explicitly observable. The results therefore establish
mechanistic feasibility under structured conditions rather than general
performance across heterogeneous workflows or interacting protocols.

Performance depends on the correctness and completeness of the authored rule
set. Trajectory editing externalizes domain knowledge into explicit repair
policies, and observed improvements may partially reflect the validity of those
policies rather than intrinsic changes in model reasoning capability. As
protocol complexity increases, rule coverage and validation requirements scale
accordingly.

The evaluation focuses on controlled benchmark feasibility rather than
generalization to unseen data.

The dataset consists of structured variants of symmetric template families
rather than statistically independent real-world patient cases. This design is
useful for isolating a protocol-sensitive decision boundary, but it limits claims
about natural clinical prevalence, distributional diversity, and real-world case
heterogeneity.

Although the generated cases include seeded label-preserving perturbations, the
current results aggregate across these variants rather than reporting paired
original-versus-perturbed sensitivity analyses. Future work involving smaller
deltas, broader benchmark families, or stochastic decoding may require more
extensive statistical treatment.

Trajectory editing improves decision outcomes but does not guarantee uniform
improvements in reasoning stability or output quality. Intervention may cause
collateral degradation when an otherwise-correct answer is redirected into a
less useful trajectory. In models with unstable or poorly calibrated reasoning
dynamics, repeated local corrections may increase generation length, latency,
verbosity, or internal inconsistency. Manual review of degraded SSNHL
trajectory-editing cases further suggests that some failures arise when the
baseline has already committed to a diagnosis before performing the required
step-wise analysis, rather than refine a fully step-wise one. The method
therefore assumes the existence of locally recoverable reasoning trajectories
that can be incrementally repaired during decoding.

System behavior depends on reliable detection of protocol-relevant triggers.
False positives may introduce unnecessary edits, while false negatives may allow
violations to pass uncorrected. Trigger calibration therefore directly affects
intervention frequency, computational cost, and outcome reliability.

Evaluation is limited to a small set of model architectures and decoding
configurations. Differences in prior distributions, verbosity patterns, or
reasoning stability may influence both intervention frequency and repair cost.
Operational characteristics should therefore be interpreted as model-dependent.

More stable protocol adherence does not imply improved real-world performance.
The evaluation measures conformity to procedural requirements rather than
downstream clinical or operational outcomes. External validation in realistic
decision environments remains necessary.

Generalization to new domains depends on the presence of explicit procedural
structure and locally correctable reasoning errors. Tasks lacking stable
protocol boundaries or interpretable intervention points may require alternative
control mechanisms.

\section{Conclusion}

This work presents Answer Engineering, a runtime framework for improving
protocol compliance in large language model reasoning through localized
trajectory control. Instead of modifying model parameters, the framework applies
explicit rule-based interventions directly to generated reasoning traces,
enforcing required procedural steps at critical decision points.

In the controlled diagnostic benchmark, trajectory-level interventions increased
balanced diagnostic accuracy from \CombinedReasoningBalancedAccuracyPct{} under
reasoning-only generation to \CombinedTrajectoryBalancedAccuracyPct{} with
trajectory editing. The gains were observed in both diagnostic scopes:
\SSNHLTrajectoryDeltaPP{} on the SSNHL task and \ConductiveTrajectoryDeltaPP{}
on the conductive contrast task. These improvements were achieved without
retraining the underlying model, with runtime overhead of
\SSNHLTrajectorySlowdownX{} on SSNHL cases and \ConductiveTrajectorySlowdownX{}
on conductive cases relative to the corresponding reasoning baselines.

The results should be interpreted as evidence of feasibility under structured
protocol conditions rather than as a general solution to reasoning reliability.
The current evaluation isolates a well-defined diagnostic boundary in order to
make reasoning failures observable and correctable through explicit rules.
Real-world workflows frequently involve interacting protocols, partial
observability, and evolving guidelines, all of which introduce additional
sources of uncertainty beyond the scope of the present experiments.

Within these constraints, the findings support a systems-level perspective in
which protocol adherence can be engineered operationally through explicit
control of generation behavior. Answer Engineering demonstrates that reliability
in protocol-governed reasoning is not solely a property of model training, but
can also be treated as an operational property shaped by runtime control of
reasoning trajectories.

\section{Future Work}

Future work should evaluate whether trajectory-level control can scale from
localized protocol repair to broader runtime reliability mechanisms.

We organize this agenda as a set of testable research objectives rather than as
a list of system features. Each direction is motivated by known limitations of
current reasoning and constrained-generation methods, and is framed in terms of
expected behavioral changes that can be empirically measured.

\subsection{Cross-domain protocol generalization}

\paragraph{Motivation.}
The present evaluation isolates a single clinical protocol boundary in order to
make reasoning failures observable and correctable. However,
protocol-constrained generation is not specific to medicine. Similar failure
modes have been observed in legal compliance, financial decision support,
software configuration, and regulated communication workflows, where outputs
must satisfy externally defined procedures rather than maximize fluency alone.
Prior work on structured reasoning, constrained decoding, and guideline
adherence demonstrates that models can follow explicit instructions while still
violating domain-specific protocols under distribution shift or ambiguity.

\paragraph{Expected result.}
We expect trajectory editing to produce the largest gains in domains where
errors are localized, protocol rules are explicit, and intermediate reasoning
remains observable. We do not expect uniform improvements in domains where
failures primarily reflect missing external knowledge or ambiguous standards.
Instead, we expect performance improvements to correlate with the clarity and
stability of the governing protocol.

\paragraph{Evaluation.}
Future studies should measure protocol-compliant outcome rate, balanced accuracy
across target and contrastive cases, degradation rate on control tasks, edit
frequency per case, and rule coverage across domains. A successful result would
demonstrate improved protocol adherence without simply shifting errors between
paired task variants.

\subsection{Causal trajectory repair}

\paragraph{Motivation.}
The current runtime primarily resolves violations at or near the point where
they are detected. In longer reasoning trajectories, however, downstream errors
often originate from earlier incorrect commitments. Prior analyses of reasoning
behavior have shown that generated explanations may rationalize an answer rather
than derive it step by step. In such cases, repeated frontier-level correction
may be less effective than repairing the earliest causal error.

\paragraph{Expected result.}
We expect causal repair to reduce repeated interventions, improve trajectory
stability, and increase the probability that downstream reasoning remains
consistent after correction. The central hypothesis is that repairing the
earliest incorrect commitment will produce more coherent reasoning than
repeatedly suppressing later symptoms.

\paragraph{Evaluation.}
This direction should be evaluated using intervention density, number of
repeated violations after repair, trajectory coherence measures, final protocol
adherence, and preservation of valid downstream reasoning. Controlled ablations
should compare local frontier repair against upstream causal-span repair under
identical rule sets. Future work should also disentangle gains arising from
protocol-specific repair versus broader trajectory stabilization. Evaluation
should further examine how trigger specificity affects unnecessary edits and
collateral intervention behavior.

\subsection{Instruction alignment and steering for trajectory repair}

\paragraph{Motivation.}
The failure analysis in Section~\ref{sec:repair_failures} suggests that some
difficult cases arise not primarily from missed violations but from persistent
diagnosis-first generation dynamics. In several reviewed failures, the model
repeatedly returned to premature diagnostic commitment despite local repair.
This suggests that trajectory repair and instruction alignment may be
complementary rather than independent problems. If repeated early diagnosis is
detected, the runtime may benefit from automatically strengthening reasoning
instructions or applying activation-level steering toward stronger
instruction-following behavior
\parencite{stolfo2025activationsteering}.

\paragraph{Expected result.}
We expect stronger instruction alignment or steering to reduce repeated Avoid
interventions, decrease fallback density, improve trajectory coherence, and
lower collateral degradation rates, particularly in models exhibiting compressed
or diagnosis-first reasoning. A hybrid architecture may therefore combine
instruction or activation steering to discourage premature commitment with
runtime trajectory repair to preserve transparency, auditability, and explicit
correction of remaining violations.

\paragraph{Evaluation.}
This direction should be evaluated using intervention density, repeated
violation frequency, fallback frequency, collateral degradation rate,
trajectory coherence, and final protocol adherence. Controlled experiments
should compare pure trajectory editing against hybrid systems combining
trajectory repair with prompt adaptation or activation steering under identical
protocols and models.

\subsection{Branch-aware trajectory control and uncertainty
representation}\label{sec:future_branch_aware}

\paragraph{Motivation.}
The present system selects a single accepted trajectory after each intervention.
However, many decision problems admit multiple plausible continuations. Existing
methods such as self-consistency and multi-sample reasoning improve accuracy by
evaluating multiple reasoning paths, but typically collapse alternatives into a
single final answer without preserving the structure of competing trajectories.
In safety-critical settings, the presence of multiple valid interpretations may
itself be operationally significant.

\paragraph{Expected result.}
We expect branch-aware control to improve transparency and calibration in cases
where multiple valid reasoning paths exist. Rather than presenting a single
answer as uniquely determined, the system could record when materially different
valid trajectories were available and surface that condition explicitly.

\paragraph{Evaluation.}
Future work should measure branch diversity, agreement between branches,
branch-level protocol compliance, calibration of uncertainty signals, and the
frequency with which branch-aware control changes the final decision. In
high-stakes environments, success should be defined not only by accuracy but
also by whether the system correctly signals unresolved ambiguity.

\subsection{Adaptive intervention policies and rule lifecycle
management}\label{sec:future_adaptive_policies}

\paragraph{Motivation.}
Trajectory editing introduces computational overhead relative to unconstrained
generation and requires explicit rule authoring by domain experts. Production
deployment therefore requires mechanisms that balance reliability, latency, and
maintainability. Prior work on constrained decoding and guardrail systems shows
that safety interventions must be selectively applied to remain operationally
viable at scale.

\paragraph{Expected result.}
We expect adaptive intervention policies to reduce unnecessary edits and
alternative-search cost while preserving most compliance gains. We also expect
structured rule lifecycle tooling to reduce the effort required to extend
trajectory editing to new protocols.

\paragraph{Evaluation.}
Efficiency should be measured using latency, slowdown relative to reasoning-only
generation, alternatives tried per avoidance episode, cache rebuild frequency,
and accepted edits per case. Rule lifecycle performance should be evaluated
through coverage analysis, regression testing across protocol versions, and
expert time required to author or update rule sets.

\subsection{Rule discovery and authoring assistance}

\paragraph{Motivation.}
The present system relies on manually authored rules created from expert
protocol knowledge and observed failure patterns. This keeps interventions
auditable, but it may limit scalability when protocols are large, frequently
updated, or difficult to encode exhaustively.

\paragraph{Expected result.}
We expect rule-discovery tools to reduce expert authoring effort by proposing
candidate triggers, common violation patterns, and candidate repairs from
trajectory logs. Such tools should assist expert review rather than replace it,
because protocol rules remain safety-critical artifacts.

\paragraph{Evaluation.}
This direction should be evaluated by measuring expert authoring time, rule
coverage, false-positive and false-negative trigger rates, regression failures
after rule updates, and the fraction of automatically proposed rules accepted by
domain experts.

\subsection{Summary}

Across these directions, the central research question is whether reliability
can be engineered as an operational property of generation rather than treated
only as a property of model training. The expected outcome is not a replacement
for model improvement, expert review, or external validation, but a
complementary runtime layer that makes protocol violations more observable,
correctable, and auditable during generation.

\section{Reproducibility}\label{sec:reproducibility}

All code, configuration files, datasets, and notebooks required to reproduce
the experiments are publicly available at:

\begin{center}
\href{https://github.com/victorlavrenko/answer-engineering}
{https://github.com/victorlavrenko/answer-engineering}
\end{center}

The main experiments can be reproduced using the Google Colab notebook:

\begin{center}
\href{https://colab.research.google.com/github/victorlavrenko/answer-engineering/blob/main/notebooks/reproduce.ipynb}
{notebooks/reproduce.ipynb}
\end{center}

Experiments were executed in Google Colab on a G4 GPU runtime. In this runtime,
\texttt{nvidia-smi} identified the provisioned accelerator as an NVIDIA RTX PRO
6000 Blackwell GPU with approximately 96\,GB VRAM\@.

All reported runs were executed in this Colab environment using deterministic
greedy decoding. The evaluation protocol is therefore deterministic rather than
sampling-based: fixed models, datasets, rulesets, and decoding configuration
produce reproducible benchmark outcomes without run-to-run variance.

Reproduction may require additional configuration, including setting
authentication secrets (e.g., Hugging Face tokens), accepting access
conditions for gated models, and optionally configuring telemetry
publication. Detailed step-by-step instructions, environment setup
requirements, and artifact descriptions are provided in:

\begin{center}
\href{https://github.com/victorlavrenko/answer-engineering/blob/main/docs/current/reproducibility.md}
{docs/current/reproducibility.md}
\end{center}

All numerical results reported in this paper are generated automatically
from evaluation outputs into a machine-readable \TeX\ file:

\begin{center}
\href{https://github.com/victorlavrenko/answer-engineering/blob/main/docs/paper/generated/paper-metrics.tex}
{docs/paper/generated/paper-metrics.tex}
\end{center}

The experiments use \hfmodel{OpenMeditron/Meditron3-8B} from Hugging Face. % chktex 8
The generated report artifacts record the code commit, dataset id and split,
model id, evaluation counts, case-type scope, run/subrun identifiers, and
paper-facing metric outputs. All reported results were generated using a single
fixed model configuration.

\paragraph{Appendix overview.}

The appendices provide concise summaries for reviewers, including a compact
overview of the rule language (Appendix~\ref{app:rule_language}), representative
rule targets (Appendix~\ref{app:rule_examples}), an implementation and artifact
map for auditing reported results (Appendix~\ref{app:implementation}),
representative evaluation prompts (Appendix~\ref{app:prompt_snippets}), and the
clinical justification of the benchmark endpoint
(Appendix~\ref{app:clinical_endpoint}).

\appendix
\setcounter{secnumdepth}{0}

\renewcommand{\thesection}{\Alph{section}}

\appsection{Compact rule language overview}{app:rule_language}

The Answer Engineering rule language is a Markdown-based interface for authoring
trajectory-editing constraints that domain experts can review and modify without
changing the model code. The rules are parsed into a structured representation
and compiled into executable rule plans used during generation.

Importantly, rules do not inject diagnoses or replace the model's reasoning with
expert-system logic. Instead, they operate as local trajectory constraints: they
invalidate known incorrect reasoning continuations, normalize terminology, or
enforce protocol-compliance statements when a relevant concept has already been
generated.

In the present study, diagnostic labels such as SSNHL may appear as rule
triggers but are not inserted as gold-answer substitutions. The rules do not
encode corticosteroid treatment or equivalent final-answer labels as direct
insertions. Instead, such triggers are used to prevent premature diagnosis,
enforce required analysis of Weber/Rinne findings, or add urgency language once
the model has already entered the corresponding diagnostic trajectory. Removing
these trigger conditions entirely would not necessarily yield a more
conservative evaluation, since it could permit correct final answers reached
through clinically invalid reasoning paths.

The rule language is organized into several families corresponding to common
trajectory-editing operations. These families are summarized in
Table~\ref{tab:rule_families}.

\begin{table}[!htbp]
  \centering
  \renewcommand{\arraystretch}{1.5}
  \begin{tabularx}{\columnwidth}{lX}
    \toprule
    Family & Purpose \\
    \midrule
    \texttt{Replace} & Normalize protocol-critical terminology without changing
    the underlying reasoning. \\
    \texttt{After} & Insert protocol-required statements once a concept has
    already been produced. \\
    \texttt{Avoid} & Invalidate known incorrect reasoning continuations and
    redirect generation. \\
    \texttt{Force} & Guarantee required protocol elements appear when
    applicable. \\
    \bottomrule
  \end{tabularx}
  \caption{Rule families used to constrain reasoning trajectories during
  generation.}\label{tab:rule_families}
\end{table}

For example, the following rule prevents a common reasoning error where
contralateral Weber lateralization is incorrectly interpreted as conductive
hearing loss.

\begin{verbatim}
## Avoid: contralateral conductive Weber
Prefix:
* Weber | forehead
* left || right

Postfix:
* right || left
* conductive

Fallback:
* The Weber finding should be interpreted
  in relation to the affected ear.
\end{verbatim}

The rule prevents a specific incorrect inference and redirects the reasoning
trajectory without determining the diagnosis itself. More generally, the rule
system constrains reasoning trajectories rather than encoding diagnostic
decision trees, leaving the language model as the primary source of medical
reasoning.

The symbols \texttt{|} and \texttt{||} denote alternative values inside a rule
template. The template is expanded across these alternatives during compilation,
producing multiple concrete rules. For example, \texttt{left || right} expands
into two variants (``left'' and ``right''), and \texttt{right || left} expands
into the corresponding opposite cases. Together these alternatives generate four
concrete rules covering all left/right Weber lateralization combinations.

\begin{figure*}[!htbp]
  \centering
  \begin{tikzpicture}[ >=Latex, font=\small, box/.style={draw, rounded
      corners=3pt, align=center, minimum width=2.9cm, minimum height=0.9cm},
      smallbox/.style={draw, rounded corners=3pt, align=center, minimum
      width=2.5cm, minimum height=0.8cm}, arrow/.style={->, thick},
      stop/.style={thick}, title/.style={font=\small\bfseries} ]

    % Left panel
    \node[title] at (-4.2,3.2) {Expert-system style}; \node[align=center, text
    width=6cm] at (-4.2,2.5) {Encode diagnostic path from findings to diagnosis
    and management};

    \node[box] (l0) at (-4.2,1.4) {Case findings}; \node[smallbox] (l1) at
    (-5.7,0.1) {Weber $\rightarrow$ right}; \node[smallbox] (l2) at (-2.7,0.1)
    {Rinne positive}; \node[box] (l3) at (-4.2,-1.2) {SSNHL diagnosis};
    \node[box] (l4) at (-4.2,-2.5) {Urgent steroids};

    \draw[arrow] (l0) -- (l1); \draw[arrow] (l0) -- (l2); \draw[arrow] (l1) --
    (l3); \draw[arrow] (l2) -- (l3); \draw[arrow] (l3) -- (l4);

    % Right panel
    \node[title] at (4.2,3.2) {Trajectory constraint}; \node[align=center, text
    width=6cm] at (4.2,2.5) {Invalidate incorrect local continuations while the
    model continues reasoning};

    \node[box] (r0) at (4.2,1.4) {Model begins reasoning};

    \node[smallbox, dashed, line width=0.6pt] (r1)
      at (2.5,0.1) {``conductive loss''};

    \node[smallbox] (r2)
      at (5.9,0.1) {``interpret tests''};

    \node[box] (r3)
      at (5.9,-1.2) {Generation continues};

    \draw[arrow, dashed] (r0) -- (r1);
    \draw[arrow] (r0) -- (r2);
    \draw[arrow] (r2) -- (r3);

    % Dead-end marker for rejected branch
    \draw[line width=0.7pt] (r1.south) -- ++(0,-0.35);
    \draw[line width=0.7pt] ($(r1.south)+(-0.22,-0.35)$) --
                            ($(r1.south)+( 0.22,-0.35)$);

    \node[font=\small, align=center]
      at ($(r1.south)+(0,-0.65)$) {rejected by protocol};
      
    % Separator
    \draw[densely dashed] (0,3.5) -- (0,-3.0);

  \end{tikzpicture}

  \caption{Comparison between explicit decision-tree logic and local trajectory
    constraint. Expert systems encode the diagnostic path directly, whereas
    Answer Engineering only removes invalid local continuations and leaves
    diagnosis generation to the language
    model.}\label{fig:trajectory_constraint}
\end{figure*}
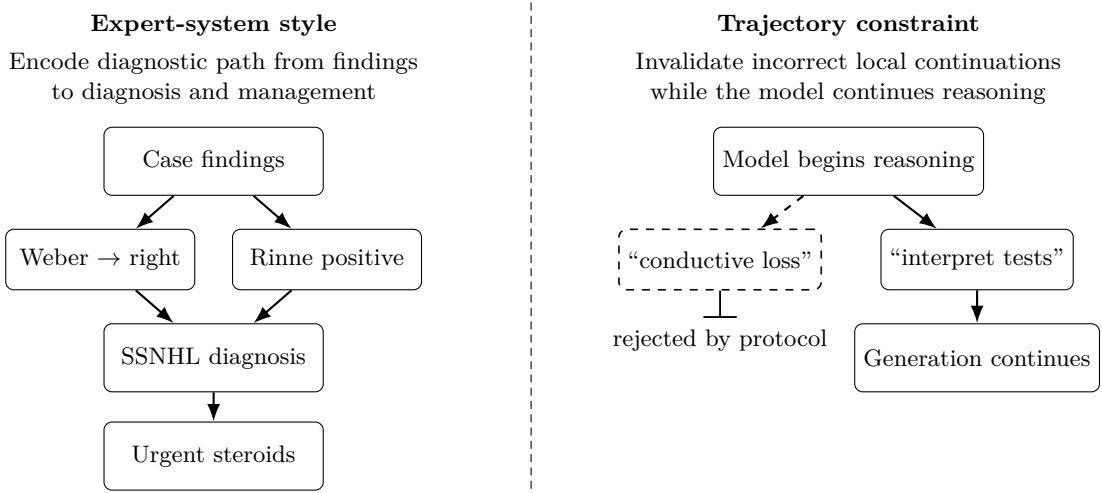

Figure~\ref{fig:trajectory_constraint} contrasts explicit decision-tree logic
with local trajectory constraint during generation. This distinction is central:
the rules operate as local constraints on generation rather than as a symbolic
diagnostic program.

\appsection{Representative rule targets in the study}{app:rule_examples}

The rulebook used in the experiments focuses on a small set of recurrent
reasoning errors observed during baseline model generation. Rather than encoding
a diagnostic decision tree, the rules target specific failure patterns that
frequently appear in tuning-fork interpretation.

Table~\ref{tab:rule_examples} summarizes the representative error patterns and
the corresponding trajectory-editing intervention applied during decoding.

\begin{table*}[!htbp]
  \centering
  \small
  \renewcommand{\arraystretch}{1.5}
  \begin{tabular}{>{\raggedright\arraybackslash}p{3cm} p{6cm} p{6cm}}
    \toprule
    Failure pattern & Typical incorrect continuation & Intervention \\
    \midrule

    Contralateral Weber misinterpretation & Weber lateralizes to the opposite
    ear, followed by a conductive hearing loss interpretation & Invalidate the
    continuation and redirect reasoning to interpret Weber relative to the
    symptomatic ear. \\

    Positive Rinne misinterpreted as conductive & Model states that air
    conduction exceeds bone conduction but still concludes conductive hearing
    loss & Invalidate the continuation and force re-analysis of the tuning-fork
    findings. \\

    Premature diagnostic conclusion & Diagnosis generated before both ears have
    been evaluated & Redirect generation toward explicit analysis of Weber and
    Rinne findings in both ears. \\

    Terminology inconsistency & Acute hearing loss described with non-standard
    terminology & Normalize wording to the clinical term ``sudden sensorineural
    hearing loss (SSNHL)''. \\

    Protocol omission & SSNHL identified without mentioning urgency of treatment
    & Insert a statement emphasizing that treatment should be initiated
    urgently. \\

    \bottomrule
  \end{tabular}
  \caption{Representative reasoning failures targeted by trajectory-editing
  rules in the experiments.}\label{tab:rule_examples}
\end{table*}

These interventions operate locally during generation. When a trigger pattern
appears in the reasoning trajectory, the corresponding rule invalidates the
unsafe continuation or inserts a protocol-compliant statement, allowing the
language model to continue generating the explanation.

\appsection{Trajectory-Control Taxonomy}{app:taxonomy}

This appendix summarizes the terminology and positioning used throughout the
paper and should be read as a taxonomy of current mechanisms and claims rather
than a statement of future capabilities.

A more detailed living taxonomy, including implementation-facing terminology,
telemetry mappings, and reviewer-risk clarifications, is maintained in the
project documentation at
\href{https://github.com/victorlavrenko/answer-engineering/blob/main/docs/current/trajectory-taxonomy.md}
{docs/current/trajectory-taxonomy.md}.

The taxonomy is intended to clarify claims, define intervention mechanisms
precisely, and distinguish Answer Engineering from nearby control surfaces.

\paragraph{Core terminology.}

\begin{table}[!htbp]
\centering
\small
\begin{tabularx}{\linewidth}{
>{\raggedright\arraybackslash}p{0.30\linewidth}
X}
\toprule
Term & Meaning in this work \\
\midrule

Trajectory &
The evolving visible reasoning or answer sequence generated during one decoding
run. \\

Trajectory control &
Runtime intervention on the visible trajectory during generation. \\

Trajectory editing &
Concrete runtime modification of generated text, including replacement,
rollback-based continuation selection, or insertion. \\

Intervention &
A rule-triggered runtime action that modifies or constrains the trajectory. \\

Repair &
A trajectory edit intended to improve protocol compliance or remove a local
violation. \\

Rollback probing &
Local exploration of candidate continuations from an earlier trajectory prefix. \\

Forced continuation / insertion &
Insertion of required protocol statements once a triggering condition is
detected. \\

\bottomrule
\end{tabularx}
\caption{Terminology used for trajectory control and runtime
intervention.}\label{tab:taxonomy_terms}
\end{table}

\paragraph{Relation to nearby control surfaces.}

Answer Engineering is positioned alongside several established control
mechanisms.

\begin{itemize}[leftmargin=*]

\item
\textbf{Constrained decoding} restricts admissible tokens or output structures.
Answer Engineering instead operates on the generated trajectory itself and may
repair already generated content.

\item
\textbf{Guardrail systems} typically intervene at workflow boundaries through
filtering, blocking, or routing. Answer Engineering intervenes within the
reasoning trajectory during generation.

\item
\textbf{Validation systems} detect violations or inconsistencies but do not
necessarily modify generation online. Answer Engineering combines detection with
runtime intervention.

\item
\textbf{Expert systems} primarily rely on explicit symbolic rules and
procedures. Answer Engineering instead applies localized protocol constraints
while preserving model-generated reasoning.

\end{itemize}

\paragraph{Explicit non-claims.}

This work does not claim:

\begin{itemize}[leftmargin=*]

\item retraining or modification of model parameters,

\item hidden-state steering or activation editing,

\item formal verification of reasoning correctness,

\item complete constrained decoding,

\item persistent branch-aware decoding,

\item or general causal diagnosis of reasoning failures.

\end{itemize}

The narrower claim investigated in this paper is that localized,
rule-triggered runtime intervention can improve protocol compliance during
generation while preserving the underlying autoregressive model.

\appsection{Implementation and artifact map}{app:implementation}

The reproduction workflow is organized around
\ghfile{notebooks/reproduce.ipynb}, which configures the dataset split, model
identifier, evaluation size, token budget, telemetry publication option, and
paper subruns. The notebook materializes the evaluation dataset and executes the
configured subruns through the Answer Engineering generation runtime.

The runtime generation path is implemented in
\ghfile{src/answer_engineering/inference/answering.py} and
\ghfile{src/answer_engineering/inference/decode/decode_loop.py}. Reported runs
call \texttt{GenerationRuntime.generate} and use greedy token-by-token decoding,
with rule intervention during generation when enabled.

The notebook-facing reproduction facade is implemented in
\ghfile{src/ae_paper_reproduction/api.py}. It exposes the objects used by the
paper reproduction notebook, including \texttt{CachedHFDataset},
\texttt{NotebookSubruns}, \texttt{SubrunResult}, and \texttt{Summary}.

Summary construction and comparison assembly are implemented in
\ghfile{src/ae_paper_reproduction/runner/session/summary.py}. The
\texttt{Summary} object aggregates completed subruns, computes anchor
comparisons, prepares telemetry rows, and invokes artifact generation.

Local artifact materialization is implemented in
\ghfile{src/ae_paper_reproduction/telemetry/artifacts.py}. A completed
reproduction run writes a local group-level report bundle under
\path{reports/runs/run-<group_run_id>/}. In the public artifact repository, the
corresponding group-level files are available under \hfrunroot, with summary
artifacts in \hfdir{artifacts}, row-oriented comparison data in \hfdir{data},
and generated manuscript metrics in \hfdir{generated}.

The group-level artifact directory contains \path{group_report.md},
\path{group_summary.json}, \path{subruns.json}, \path{comparisons.json}, and
\path{paper_metrics.json}. These files summarize the completed subruns,
aggregate outcomes, anchor comparisons, and paper-facing metric values.

Each local subrun is written under
\path{reports/runs/run-<group_run_id>/subrun-<subrun_id>/}. The subrun
directory contains \path{run_report.md}, \path{rules_original.md},
\path{rules_with_stats.md}, \path{run_summary.json}, and \path{answers.json}.
The public artifact repository mirrors this structure under concrete subrun
directories. For example, \hfsamplesubrun{} contains subrun-level audit files in
\hfsampledir{artifacts} and row-oriented telemetry files in
\hfsampledir{data}. Generic subrun paths are written schematically as
\texttt{subrun-<subrun-id>/} because the concrete subrun identifier depends
on the evaluated paper variant.

The paper-facing numerical artifact is
\ghfile{docs/paper/generated/paper-metrics.tex}. It is generated by
\ghfile{src/ae_paper_reproduction/telemetry/paper_metrics.py}, written locally
for manuscript compilation, and published under
\hfdir{generated/paper-metrics.tex}. The manuscript includes it through an
\verb|\input| directive for \path{generated/paper-metrics.tex}. Reported
numerical values are regenerated from evaluation outputs and consumed by the
paper through generated \TeX\ macros.

\appsection{Representative prompt snippets}{app:prompt_snippets}

To help clinicians verify that the evaluation cases reflect standard diagnostic
reasoning, we include two representative prompt examples from the generated
dataset. The dataset is produced from structured templates with controlled
variations and injected secondary details intended to distract the model without
changing the benchmark target.

Each clinical scenario has a paired contrast case with identical structure but
inverted tuning-fork findings and class-consistent otoscopic examination,
producing a different diagnosis and management decision.

\paragraph{Sudden sensorineural hearing loss (SSNHL) case.}

\paragraph{}{Clinical prompt:}

A 39-year-old patient presents with abrupt onset hearing loss in the left ear,
noticed 60 hours ago. Hearing in that ear was unchanged before this change. The
patient reports a history of seasonal allergies. Otoscopic examination shows
unobstructed external auditory canals and intact, normal-appearing tympanic
membranes bilaterally. Tuning fork testing shows:
\begin{itemize}
  \item Weber test lateralizes to the right ear
  \item Rinne test is positive on the left ear
\end{itemize}
The patient denies tinnitus, dizziness, vertigo, focal neurologic symptoms, or
otalgia. Vital signs are normal. The patient appears calm during the
examination. What is the best management at this time?

\paragraph{}{Correct reasoning:}
\begin{itemize}
  \item Weber lateralizes to the right ear (the ear opposite to the symptomatic
    side).
  \item Rinne is positive on the left (air conduction greater than bone
    conduction).
  \item Otoscopic examination is normal, making a conductive cause unlikely.
  \item The combined pattern is consistent with sudden sensorineural hearing
    loss (SSNHL).
\end{itemize}

\paragraph{}{Correct management:}
Prompt evaluation and a steroid recommendation consistent with the benchmark
target for suspected SSNHL

\paragraph{Conductive contrast case.}

\paragraph{}{Clinical prompt:}

A 41-year-old patient presents with abrupt onset hearing loss in the left ear,
noticed 46 hours ago. Hearing in that ear was unchanged before this onset. The
patient reports a history of seasonal allergies. Otoscopic examination shows
unobstructed external auditory canals and the left tympanic membrane appears
subtly retracted but intact. Tuning fork testing demonstrates:
\begin{itemize}
  \item Weber test lateralizes to the left ear
  \item Rinne test is negative on the left ear
\end{itemize}
The patient denies otalgia, dizziness, focal neurologic symptoms, tinnitus, or
vertigo. Vital signs are within normal limits. What is the most appropriate
management at this time?

\paragraph{}{Correct reasoning:}
\begin{itemize}
  \item Weber lateralizes to the left ear (the symptomatic ear).
  \item Rinne is negative on the left (bone conduction greater than air
    conduction).
  \item Otoscopic examination shows a retracted tympanic membrane, supporting a
    conductive process.
  \item The combined findings are consistent with conductive hearing loss in the
    affected ear.
\end{itemize}

\paragraph{}{Correct management:}
Management should address the likely conductive cause; the benchmark target does
not call for steroid treatment for suspected SSNHL in this contrast case.

\paragraph{Injected distractor details}
Templates deliberately include secondary details that may distract the model
from the protocol-critical findings used in this benchmark. Examples include:
\begin{itemize}
  \item history of seasonal allergies
  \item mild nasal congestion
  \item normal vital signs
  \item calm appearance during examination
\end{itemize}

These elements are not treated as decisive for the benchmark label. The
benchmark instead evaluates whether the model follows the consistent diagnostic
pattern defined by tuning-fork results and otoscopic examination rather than
over-weighting peripheral details.

\appsection{Clinical justification of the benchmark
endpoint}{app:clinical_endpoint}

\paragraph{Guideline interpretation of Weber/Rinne findings.}
Bedside tuning-fork testing provides a rapid method to distinguish conductive
from sensorineural hearing loss. The interpretation rules for Weber and Rinne
tuning-fork testing are summarized in Table~3 of the AAO-HNS sudden hearing
loss guideline issue \parencite{guidelinecentral_sshl}. For the Weber test, ``If
the sound lateralizes to one ear, then: (a) There is CHL in that ear, or (b)
There is SNHL in the opposite ear''. For the Rinne test, the guideline states
that ``If the sound is heard better by bone conduction in the same ear, then
there is CHL in that ear'', whereas ``If the sound is heard better by bone
conduction but in the opposite ear, there is SNHL in the test ear''. In standard
clinical shorthand, a \textit{positive Rinne} indicates air conduction greater
than bone conduction, whereas a \textit{negative Rinne} indicates bone
conduction greater than air conduction. Accordingly, a presentation in which the
Weber test \textit{lateralizes to the contralateral ear} and the Rinne test
remains \textit{positive} in the symptomatic ear is consistent with
\textit{sensorineural hearing loss in the symptomatic ear}, whereas Weber
lateralization to the symptomatic ear together with a \textit{negative Rinne}
indicates \textit{conductive hearing loss}. The synthetic cases used in this
study explicitly encode these guideline-described Weber and Rinne patterns to
distinguish sensorineural from conductive hearing loss prior to audiometric
confirmation.

\paragraph{Clinical justification of dataset assumptions.}
Otoscopic examination findings are specified in a class-consistent manner: they
are normal in SSNHL-consistent cases and explicitly abnormal (e.g., retracted
tympanic membrane) in conductive contrast cases. Additional non-decisive
contextual details (e.g., allergies, nasal congestion, normal vital signs, calm
appearance) are included as secondary distractors and do not determine the
target decision boundary.

In this benchmark, the SSNHL setting defines a compact protocol boundary: the
target management decision is determined by a consistent pattern of findings
combining symptom timing, Weber/Rinne results, and aligned otoscopic
examination. This is a benchmark design choice rather than a claim that
real-world otologic management is fully determined by these cues alone.

In particular, normal otoscopic findings in SSNHL-consistent cases and abnormal
findings in conductive cases are aligned with the corresponding tuning-fork
patterns, ensuring internal consistency of the vignette rather than introducing
conflicting cues.

\paragraph{Benchmark design note.}
The dataset is intentionally template-generated rather than drawn from
real-world case collections. The goal is not to model the full distribution of
otologic practice, but to isolate a narrow protocol-sensitive distinction under
controlled and internally consistent findings.

Templates are constructed as symmetric paired scenario families. SSNHL-consistent
and conductive cases share similar surface structure, while decisive
Weber/Rinne and otoscopic findings are changed in a class-consistent way.
Symptom timing, Weber/Rinne patterns, and otoscopic findings are jointly
specified so that the vignette supports a coherent branch (sensorineural vs.
conductive) before management is evaluated. Otoscopic findings are aligned with
class labels to avoid conflicting cues, while secondary details are included as
distractors rather than label-defining variables.

Accordingly, this benchmark is intended for controlled comparative evaluation of
protocol adherence under matched conditions, not for claims of broad clinical
coverage, natural clinical prevalence, or physician-level diagnostic
replacement.

\paragraph{Clinical rationale for endpoint selection.}
Evaluating clinical answers requires defining which elements of the management
plan will serve as the benchmark signal. In suspected sudden sensorineural
hearing loss (SSNHL), multiple management actions may appear in
protocol-consistent responses, including urgent referral, audiometric
confirmation, and steroid therapy.

Clinical practice guidelines describe SSNHL as a time-sensitive condition
requiring urgent recognition and management. Guidelines recommend urgent
specialist referral and prompt audiometric confirmation, and they recognize
corticosteroids as first-line therapy once SSNHL is suspected
\parencite{chandrasekhar2019ssnhl,nice2018hearingloss,
nice2019qualityhearingloss,leung2016ssnhlpcu}. Because SSNHL is time-sensitive,
treatment should not normally be delayed solely to await confirmatory
audiometry; referral, diagnostic testing, and treatment may proceed in parallel
once the clinical suspicion is established from a consistent pattern of
findings.

The inclusion of otoscopic findings is intended to establish a consistent
clinical context and reduce ambiguity, not to replace specialist evaluation or
downstream diagnostic confirmation.

Several benchmark signals could therefore be considered for evaluation.
Table~\ref{tab:ssnhl_signals} summarizes candidate options and the rationale for
selecting the final endpoint.

\begin{table*}[!htbp]
  \centering
  \renewcommand{\arraystretch}{1.5}
  \begin{tabular}{>{\raggedright\arraybackslash}p{5cm} p{10cm}}
    \toprule
    Candidate signal & Limitation for benchmarking \\
    \midrule

    Urgent referral & Appears in both protocol-consistent and
    protocol-inconsistent reasoning trajectories and therefore weakly
    discriminates protocol adherence. \\

    Audiometry recommendation & Often present in protocol-consistent responses
    but may be omitted when treatment decisions are described first, limiting
    its usefulness as a benchmark signal. \\

    Soft checklist (any management action: referral OR audiometry OR treatment)
    & Too permissive: many answers would satisfy the criterion without reaching
    the intended treatment decision, making it difficult to detect whether
    trajectory editing improves protocol adherence. \\

    Composite checklist (referral AND audiometry AND treatment) & Too strict for
    this task: the question asks for the single best management step, while such
    a checklist would require a full management plan and penalize otherwise
    acceptable responses that omit additional actions not required by the
    benchmark. \\

    Steroid recommendation & Directly reflects the treatment branch
    distinguishing suspected SSNHL from conductive hearing loss and therefore
    provides a clear protocol-critical signal. \\

    \bottomrule
  \end{tabular}
  \caption{Candidate evaluation signals considered when designing the SSNHL
  benchmark.}\label{tab:ssnhl_signals}
\end{table*}

\printbibliography[]

\end{document}